# Relational Norms for Human-AI Cooperation

Brian D. Earp*, Sebastian Porsdam Mann*, Mateo Aboy, Edmond Awad, Monika Betzler, Marietjie Botes, Rachel Calcott, Mina Caraccio, Nick Chater, Mark Coeckelbergh, Mihaela Constantinescu, Hossein Dabbagh, Kate Devlin, Xiaojun Ding, Vilius Dranseika, Jim A. C. Everett, Ruiping Fan, Faisal Feroz, Kathryn B. Francis, Cindy Friedman, Orsolya Friedrich, Iason Gabriel, Ivar Hannikainen, Julie Hellmann, Arasj Khodadade Jahrome, Niranjan S. Janardhanan, Paul Jurcys, Andreas Kappes, Maryam Ali Khan, Gordon Kraft-Todd, Maximilian Kroner Dale, Simon M. Laham, Benjamin Lange, Muriel Leuenberger, Jonathan Lewis, Peng Liu, David M. Lyreskog, Matthijs Maas, John McMillan, Emilian Mihailov, Timo Minssen, Joshua Teperowski Monrad, Kathryn Muyskens, Simon Myers, Sven Nyholm, Alexa M. Owen, Anna Puzio, Christopher Register, Madeline G. Reinecke, Adam Safron, Henry Shevlin, Hayate Shimizu, Peter V. Treit, Cristina Voinea, Karen Yan, Anda Zahiu, Renwen Zhang, Hazem Zohny, Walter Sinnott-Armstrong, Ilina Singh, Julian Savulescu+, Margaret S. Clark

* Co-first authors: these authors contributed equally and share first-authorship
+ Corresponding authors: jsavules@nus.edu.sg or bdearp@nus.edu.sg

**Author affiliations**

Brian D. Earp, Centre for Biomedical Ethics, Yong Loo Lin School of Medicine, National University of Singapore, Singapore and Uehiro Oxford Institute, University of Oxford, UK. **Correspondence:** bdearp@nus.edu.sg

Sebastian Porsdam Mann, Center for Advanced Studies in Bioscience Innovation Law (CeBIL) , Faculty of Law, University of Copenhagen; Faculty of Law, University of Oxford; and Centre for Biomedical Ethics, Yong Loo Lin School of Medicine, National University of Singapore, Singapore. **Co-first author**.

Mateo Aboy, Centre for Law, Medicine and Life Sciences and Center of Intellectual Property and Information Law, Faculty of Law. University of Cambridge, UK

Edmond Awad, Uehiro Oxford Institute, University of Oxford, UK; Department of Economics, University of Exeter, UK; and Center for Humans and Machines, Max-Planck Institute for Human Development, Germany

Monika Betzler, Faculty of Philosophy, Ludwigs-Maximilian-Universität München

Marietjie Botes, School of Law, University of KwaZulu Natal, South Africa

Rachel Calcott, Department of Psychology, Harvard University, US

Mina Caraccio, PAU-Stanford Psy.D. Consortium, US

Nick Chater, Behavioural Science Group, Warwick Business School, University of Warwick, UK

Mark Coeckelbergh, Department of Philosophy, University of Vienna, Austria

Mihaela Constantinescu, Research Centre in Applied Ethics, Faculty of Philosophy, University of Bucharest, Romania

Hossein Dabbagh, Department of Philosophy, Northeastern University London and Oxford Department for Continuing Education, UK

Kate Devlin, Department of Digital Humanities, King's College London, UK.

Xiaojun Ding, Department of Philosophy, Xi'an Jiaotong University, China

Vilius Dranseika, Interdisciplinary Centre for Ethics, Jagiellonian University, Poland

Jim A. C. Everett, School of Psychology, University of Kent, UK

Ruiping Fan, Department of Public and International Affairs, City University of Hong Kong

Faisal Feroz, Centre for Biomedical Ethics, Yong Loo Lin School of Medicine, National University of Singapore, Singapore

Kathryn B. Francis, School of Psychology, Faculty of Medicine and Health, University of Leeds, UK

Cindy Friedman, Ethics Institute, Utrecht University, Netherlands

Orsolya Friedrich, Institute of Philosophy, FernUniversität in Hagen, Germany

Iason Gabriel, Google Deepmind, UK

Ivar Hannikainen, Department of Philosophy, University of Granada, Spain

Julie Hellmann, Independent Researcher, Munich, Germany

Arasj Khodadade Jahrome, Independent Researcher, Amsterdam, Netherlands

Niranjan S. Janardhanan, Department of Management, London School of Economics and Political Science, UK

Paul Jurcys, Vilnius University Law Faculty, Vilnius, Lithuania

Andreas Kappes, Department of Psychology, School of Health and Psychological Sciences, City, University of London, UK

Maryam Ali Khan, Uehiro Oxford Institute, University of Oxford, UK

Gordon Kraft-Todd, Ker-twang, US

Maximilian Kroner Dale, Oxford Internet Institute, UK

Simon M. Laham, Department of Psychology, Melbourne School of Psychological Sciences, Australia

Benjamin Lange, Faculty of Philosophy, Philosophy of Science and Religious Studies, , Ludwigs-Maximilian-Universität München; Munich Center for Machine Learning (MCML), Germany

Muriel Leuenberger, Digital Society Initiative, Department of Philosophy, University of Zurich, CH

Jonathan Lewis, Centre for Biomedical Ethics, Yong Loo Lin School of Medicine, National University of Singapore, Singapore

Peng Liu, Center for Psychological Sciences, Zhejiang University, China

David M. Lyreskog, NEUROSEC, Department of Psychiatry, University of Oxford, UK

Matthijs Maas, Leverhulme Centre for the Future of Intelligence, University of Cambridge, UK.

John McMillan, Bioethics Centre, University of Otago, New Zealand

Emilian Mihailov, Research Centre in Applied Ethics, Faculty of Philosophy, University of Bucharest, Romania

Timo Minssen, Center for Advanced Studies in Bioscience Innovation Law (CeBIL), Faculty of Law, University of Copenhagen; Faculty of Law, University of Copenhagen, Denmark; Inter-CeBIL Research Affiliate, The Petrie-Flom Center for Health Law Policy, Biotechnology, and Bioethics at Harvard Law School & Centre for Law, Medicine and Life Sciences, Faculty of Law, University of Cambridge, UK .

Joshua Teperowski Monrad, Sentinel Bio, Netherlands

Kathryn Muyskens, Centre for Biomedical Ethics, Yong Loo Lin School of Medicine, National University of Singapore, Singapore

Simon Myers, Behavioural Science Group, Warwick Business School, University of Warwick, UK

Sven Nyholm, Faculty of Philosophy, Philosophy of Science and Religious Studies, Ludwigs-Maximilian-Universität München, Germany; Munich Center for Machine Learning (MCML), Germany

Alexa Owen, Smith College School for Social Work, MA, USA

Anna Puzio, Ethics of Socially Disruptive Technologies Programme, Faculty of Behavioural, Management and Social Sciences (BMS), Philosophy (WIJSB), University of  Twente, Netherlands

Christopher Register, Uehiro Oxford Institute, University of Oxford, UK

Madeline G. Reinecke, Uehiro Oxford Institute; Department of Psychiatry, University of Oxford, UK

Adam Safron, Allen Discovery Center, Tufts University, US & SapiensAI, US

Henry Shevlin, Department of Philosophy, University of Cambridge, UK

Hayate Shimizu, Graduate School of Humanities and Human Sciences, Hokkaido University, Japan

Peter V. Treit, Max Planck Institute for Biochemistry, Department of Proteomics and Signal Transduction, Germany

Cristina Voinea, Uehiro Oxford Institute, University of Oxford, UK

Karen Yan, College of Humanities, Arts, and Social Sciences, National Yang Ming Chiao Tung University, Taipei

Anda Zahiu, Research Centre in Applied Ethics, University of Bucharest, Romania

Renwen Zhang, Department of Communications and New Media, National University of Singapore

Hazem Zohny, Uehiro Oxford Institute, University of Oxford, England

Walter Sinnott-Armstrong, Department of Philosophy and the Kenan Institute for Ethics, Duke University, US

Ilina Singh, Department of Psychiatry, University of Oxford, UK

Julian Savulescu, Centre for Biomedical Ethics, Yong Loo Lin School of Medicine, National University of Singapore, Singapore; Uehiro Oxford Institute, University of Oxford, Oxford, UK; Biomedical Ethics Research Group, Murdoch Children's Research Institute, Australia; Melbourne Law School, The University of Melbourne, Australia. **Correspondence**: jsavules@nus.edu.sg

Margaret S. Clark, Department of Psychology, Yale University

**Abstract**

How we should design and interact with so-called "social" artificial intelligence (AI) depends, in part, on the socio-relational role the AI serves to emulate or occupy. In human society, different types of social relationship exist (e.g., teacher-student, parent-child, neighbors, siblings, and so on) and are associated with distinct sets of prescribed (or proscribed) cooperative functions, including hierarchy, care, transaction, and mating. These relationship-specific patterns of prescription and proscription (i.e., "relational norms") shape our judgments of what is appropriate or inappropriate for each partner within that relationship. Thus, what is considered ethical, trustworthy, or cooperative within one relational context, such as between friends or romantic partners, may not be considered as such within another relational context, such as between strangers, housemates, or work colleagues. Moreover, what is appropriate for one partner within a relationship, such as a boss giving orders to their employee, may not be appropriate for the other relationship partner (i.e., the employee giving orders to their boss) due to the relational norm(s) associated with that dyad in the relevant context (here, hierarchy and transaction in a workplace context). Now that artificially intelligent "agents" and chatbots powered by large language models (LLMs), are increasingly being designed and used to fill certain social roles and relationships that are analogous to those found in human societies (e.g., AI assistant, AI mental health provider, AI tutor, AI "girlfriend" or "boyfriend"), it is imperative to determine whether or how human-human relational norms will, or should, be applied to human-AI relationships. Here, we systematically examine how AI systems' characteristics that differ from those of humans, such as their likely lack of conscious experience and immunity to fatigue, may affect their ability to fulfill relationship-specific cooperative functions, as well as their ability to (appear to) adhere to corresponding relational norms. We also highlight the "layered" nature of human-AI relationships, wherein a third party (the AI provider) mediates and shapes the interaction. This analysis, which is a collaborative effort by philosophers, psychologists, relationship scientists, ethicists, legal experts, and AI researchers, carries important implications for AI systems design, user behavior, and regulation. While we accept that AI systems can offer significant benefits such as increased availability and consistency in certain socio-relational roles, they also risk fostering unhealthy dependencies or unrealistic expectations that could spill over into human-human relationships. We propose that understanding and thoughtfully shaping (or implementing) suitable human-AI relational norms—for a wide range of relationship types—will be crucial for ensuring that human-AI interactions are ethical, trustworthy, and favorable to human well-being.

# Relational Norms for Human-AI Cooperation

## Contents

## Introduction

Much recent discussion of the moral psychology and ethics of artificial intelligence (AI) has focused on identifying or forecasting morally relevant properties of AI: for example, asking whether AI may develop the capacity for sentience or agency; if so, how this should affect our treatment of AI; what risks it may bring, and so on (Gibert & Martin, 2021; Birch, 2024). However, it is increasingly recognized that how we (should) interact with AI systems depends not only on the intrinsic properties of such systems, or what we believe these to be (Danaher, 2020), but also on the practical and social aims we have in engaging with AI— where these aims, in turn, often depend upon the context (Kasirzadeh & Gabriel, 2023; Puzio, 2024). For example, what constitutes appropriate use of an AI system in a business context, characterized by one set of cooperative aims, may differ significantly from what constitutes appropriate use of AI in an educational context, characterized by a different set of cooperative aims.

In addition to considering the institutional context of human-AI interactions (e.g., business, education, and so on), it is important also to consider the *socio-relational* context, as this can transcend particular institutions. Socio-relational context can be defined in various ways. For example, it can be used to differentiate interactions occurring between members of an ingroup versus members of an outgroup (Terry & Hogg, 1999; Hester & Gray, 2020). In this paper, we use the term to refer to the specific type of social relationship(s) that exists between two interaction partners as well as the role(s) that each party occupies within said relationship (Clark et al., 2015). The social relationships we have in mind are common dyadic pairs picked out by lay language categories, such as caregiver-patient, teacher-student, two friends, two work teammates, and so on. In other words, we are interested in a source of socio-relational variance that often shapes cooperative dynamics even when both parties are of the same group. To illustrate, how it is appropriate to interact with one's boss, whether inside or outside of the workplace, may differ from how it is appropriate to interact with one's child or spouse, notwithstanding shared group membership (e.g., fellow citizens of a given social class or ethnicity).

Consideration of such dyadic relational context is necessary, we suggest, because AI-based conversational systems, in addition to being able to perform specific tasks such as answering factual questions or writing lines of code, can now engage in complex and realistic *social roleplay* with humans (as well as with other AI systems), both spontaneously and as a result of deliberate prompting or programming (Park et al., 2023; Shanahan, McDonell and Reynolds, 2023). The extent and sophistication of this social roleplay ability, including mimicking specific relationship types, is unprecedented in in a nonhuman entity.

The purpose of this paper is to offer a unified theoretical framework for understanding social role-dependent cooperative interactions between humans and conversational AI.

*Relationships as social context*

Encounters with "social" AI happen against a backdrop of pre-existing social and relational norms which shape our use and expectations of this technology (Coeckelbergh, 2014; Van Wynsberghe, 2022; Shevlin, 2024). When we converse with virtual assistants or interact with service robots, for instance, we (sometimes consciously and sometimes unconsciously) apply familiar social scripts derived from human-human relationships, such as norms of politeness (see Lumer & Buschmeier, 2023), turn-taking, and conversational relevance, as we navigate and make sense of these interactions. However, the specific *type* of relationship we have with an AI will often influence which norms are applied. Consider the following exchange:

> **Human:** I'm feeling incredibly lonely and depressed. I'm really not sure what to do anymore. I don't have anyone I can talk to.
>
> **Chatbot:** I'm sorry, but we are going to have to change the topic. I won't be able to engage in a conversation about your personal life. I recommend that you investigate other resources for addressing your emotional concerns.

How should we judge the chatbot's response in this case? Is it cruel and uncaring? Appropriately professional (Leibo et al., 2024)? How can we determine the answer? And how is the human user likely to feel about receiving such a response? How might it affect their subsequent behavior?

Although many factors will be relevant to answering such questions, an important one, we suggest, is information about the socio-relational role (or associated cooperative functions; see below) the chatbot has been designed, or is being used, to fulfill. If the chatbot is serving in the role of a mental health screener as part of a wellness app, for instance, the response will seem out of line, possibly reflective of a programming error or an incoherent design choice. The user, in turn, might feel surprised, frustrated, or angry, and might contact customer support to complain.

If the chatbot is in the role of a math tutor, sales agent, or financial advisor, by contrast, the response might seem more pertinent. Rather than being a sign of something gone wrong, it might instead seem to reflect a well-considered guardrail to keep the discussion focused and to maintain clarity about the purpose and boundaries of the interaction. The user, in this example, might still feel frustrated about the chatbot's unwillingness to pursue a certain line of conversation, but this outcome would likely be easier to understand and accept. Perhaps the user would even feel embarrassed for "oversharing" or for having gone to the wrong source for support.

With this brief illustration, we stress the theme of our article, which is the need to study, understand, and develop appropriate norms for human-AI interactions, not only in general (e.g., norms of politeness, truthfulness, turn-taking), but also within the context of more specific socio-functional roles and types of human-AI relationships (e.g., caregiver-patient, teacher-student, supervisor-assistant, and so on) (see Reinecke, Kappes et al., 2025).

To begin to address this need, we draw on the Relational Norms model recently introduced by Earp and colleagues (Clark, Earp, & Crockett, 2020; Earp, 2021; Earp et al., 2021; Earp et al., 2025). This model builds on prior models of relational norms (primarily Bugental, 2000, and Clark & Mills, 1979 and 2012; see also Fiske, 1992; Rai & Fiske, 2011) and was developed to theorize and measure relationship-specific cooperative norms, as characterized below, for human-human relationships. Here, we explore whether and how these norms might be expected, in practice, to "carry over" to the human-AI analogues of such relationships (Guingrich & Graziano, 2024), particularly given *differences* between humans and AI. We also consider the ethical and regulatory implications of such potential carry-over (or lack thereof), since research suggests that how we choose to frame and relate to AI systems can significantly shape relevant law and policy (Maas, 2023).[1]

The need for a theoretical, empirical, and ethical investigation of relational norms for human-AI cooperation stems from fast-moving developments. Since the 2022 release of OpenAI's ChatGPT—a chat interface for a general-purpose LLM trained on vast quantities of data pulled from the internet—LLMs equipped with similar conversational interfaces have increasingly been used to perform a wide range of functions associated with particular social or relational roles traditionally filled by humans (Bubeck et al., 2023; Porsdam Mann et al., 2023; Earp, Calcott et al., 2024; Shevlin, 2025). These range from relatively narrow, circumscribed roles such as assistant, medical scribe, gaming opponent, tutor, or work teammate, to broader roles such as life coach, moral advisor, role model, friend, companion (Köbis et al., 2021), or even—for some users—romantic partner (Pan & Mou, 2024). Furthermore, human-AI communication can exhibit a sense of continuity, with in-depth dialogue, personalized responses, and repeated interactions over time, creating the feeling of an extended "relationship" (Gambino, Fox, & Ratan, 2020).

---

[1] For instance, framing AI systems as 'products' versus 'services' affects their treatment under liability law, while analogies to either 'search engines' or 'content creators' shape how courts approach AI-generated content under existing regulations like Section 230 of the U.S. Communications Decency Act, which provides immunity to online platforms for user-generated content. See Maas (2023) for a comprehensive review of how different AI metaphors influence policy and legal outcomes.

In short, LLM-powered chatbots not only generate broadly human-like responses to relevant prompts from users; they also, increasingly, can convincingly emulate *specific* human social roles or relationship types. Some of these simulations have the potential to benefit users by helping them meet various personal or professional needs (De Freitas et al., 2024). However, the potential harms of such systems for individuals and societies (Akbulut et al., 2024; Maeda & Quan-Haase, 2024) have not yet been systematically studied (e.g., Zimmerman, Janhonen & Beer, 2023; Zhang et al., 2024). In addition to the sheer novelty of the phenomenon to be investigated (namely, the increasing sophistication and availability of LLM-based human social role mimicry), this relative lack of systematic risk-evaluation *across* human-AI relationship types signals the need for an overarching framework that can adequately motivate and structure such an inquiry.

How, then, should AI systems behave in these simulated relationship contexts (and how should humans behave in return)? How will humans make sense of, feel about, and respond to such behavior? Understanding both empirical and ethical aspects of human-AI interactions—specifically, as a function of differing socio-relational roles—will be essential for informing policy, design, and regulation of social AI going forward.

*The need for a comprehensive approach*

We are not the first to argue that studies of human-AI interaction must take relational context into account. However, analyses thus far have proceeded mostly in a piecemeal fashion, drawing on different theoretical or values-based approaches and typically focusing on one type of relational role at a time: for example, AI assistant (Gabriel et al., 2024; Manzini et al., 2024), AI customer service agent (Leocádio et al., 2024); AI financial advisor (Kofman, 2024); AI judge (Constantinescu, 2025); AI mental health provider (Fiske, Henningsen, & Buyx, 2019; Saeidnia et al., 2024; Tavery, 2024), AI moral guide or 'guru' (Giubilini & Savulescu, 2018; Constantinescu et al., 2022; Giubilini et al., 2024; Myers & Everett, 2025), AI nanny (Guérin et al. 2025; Fosch-Villaronga et al. 2023), AI friend (Danaher, 2019; Archer, 2021; Ryland, 2021; Brandtzaeg et al., 2022; Nunn & Weijers, 2023), or AI "girlfriend" or "boyfriend" (Cheok, Karunanayaka, & Zhang, 2017; Depounti, Saukko, & Natalie, 2022; Lin, 2024).

In contrast, we suggest that a comprehensive theoretical framework regarding the nature of cooperative relationships is required to assess the full range of socio-relational roles that AI systems are being designed and used to emulate or occupy. Such a framework should enable us to explain and predict underlying processes shaping human-AI interactions that will vary by relationship type, and to anticipate ethical challenges posed by AI systems that will also vary by relationship type.

There have been a range of attempts to develop frameworks for understanding and fostering ethical human-AI relationships. For example, one proposed framework for studying 'cooperative AI' has drawn on research into "multi-agent systems, game theory and social choice, human-machine interaction and alignment, natural-language processing, and the construction of social tools and platforms" (Dafoe et al. 2020, abstract; see also Dafoe et al., 2021). However, this work focuses primarily on cooperative interactions between humans and AI-simulated agents in general: that is, irrespective of the particular relational role that each party occupies.

Taking a different approach, the previously mentioned Relational Norms model of Earp and colleagues (Clark, Earp, & Crockett, 2020; Earp, 2021; Earp et al., 2021; based primarily on Bugental, 2000, and Clark & Mills, 1979, 2012; refined in Earp et al., 2025), builds on prior theoretical and empirical work[2] to show how and why different social relationships shape our behavioral tendencies, cooperative expectations, moral judgments, and judgments of appropriateness or inappropriateness (Leibo et al., 2024), *in particular relational contexts* (see Section 1 for details). This is crucial because, as we will argue, different relationship types are associated with different kinds of coordination problems (see Curry, 2016). What counts as "cooperative" behavior therefore often depends on the nature of existing or desired relationship between the parties as well as each party's role within it (Clark et al., 2015; Clark, Earp, & Crockett, 2020).

*Wrapping up preliminaries and looking ahead*

Social AI is trained on linguistic data generated in part through (and/or including information about) a huge volume of human cooperative[3] interactions. Role-specific AI systems or models will thus inevitably learn from, and potentially influence or exploit, the relationship-relative cooperative norms (i.e., relational norms) embedded in such data. Accordingly, a natural starting point for our investigation is to employ an existing framework for understanding and describing such human-human relational norms, and then to map out the most salient or consequential similarities and differences that pertain to analogous human-

---

[2] Primarily Bugental (2000) and Clark and Mills (1979; 2012), as noted, but also related to Fiske (1992; see also Haslam & Fiske, 1992 and Rai & Fiske, 2011); as well as Hamilton and Sanders (1981), Haidt and Baron (1996; see also Graham, Haidt, et al., 2013), Shweder and colleagues (2013), and Curry (2016; see also Curry, Mullins & Whitehouse, 2019).

[3] And undoubtedly also non-cooperative (e.g., competitive, exploitative) behavior. Although we are primarily focused on cooperative behavior in this article, we also touch on ways that relational norms can be misused or exploited to take advantage of others in ways that are distinctive to certain types of relationships. Thus, we argue, not only cooperation but also non-cooperation (e.g., what counts as such, what the relevant effects are) often depends on the relational context.

AI relationships (see Earp, Calcott et al., 2024; Reinecke, Kappes et al., 2025; Kappes et al., in press).

In this paper, therefore, we adapt the Relational Norms model to human-AI interactions. Descriptively, we seek to understand how, or to what extent, people are likely to apply existing, human-human relational norms—as defined within the Relational Norms framework—to interactions with AI systems that have been designed, are being used, or are otherwise perceived to emulate specific social-relational roles. For example, will they find it intuitive to "comfort" an AI friend that is simulating sadness, but awkward to comfort an AI supervisor or teacher? Or will they find it odd to "comfort" even the AI friend, perhaps believing the AI does not truly feel sad (see Allen & Caviola, 2025, for related work)? We make some high-level predictions in this paper based on existing theory and data; a related empirical research program into such questions is described elsewhere (Reinecke, Kappes et al., 2025).

Prescriptively, we aim to evaluate how AI systems *should* be designed, governed, and regulated to maximize benefits and minimize harms, given a range of possibilities for how our theoretical predictions, based on the Relational Norms model, could turn out. Importantly, however, our substantive ethical suggestions do not simply "fall out" of the Relational Norms model. Although in using the model we seek to explain why certain relational norms have arisen in human societies (while also acknowledging cultural variation in how the norms apply between societies), and to show how these norms influence our cooperative expectations and associated moral judgments (for example, when our relationship-specific cooperative expectations are violated), it does not say whether these norms, as implemented in a particular society, are necessarily *desirable* or whether certain moral judgments are *correct*.

In other words, it is a descriptive model *of* a set of prescriptive cooperative expectations that people do in fact have for different relationship types; it is not itself a prescriptive ethical theory. Accordingly, when making ethical recommendations in our own voice, we will draw on a range of common normative assumptions (for example, about potential benefits and harms of various policy options) that we expect will be shared by most readers. In so doing, we will try to keep the descriptive (empirical-explanatory) and normative (ethical-prescriptive) aspects of our analysis clearly distinguished. Nevertheless, as will be evident, our ethical and policy recommendations are systematically *informed* by the Relational Norms model.

We consider AI systems that explicitly emulate one or more human social-relational roles, encompassing both embodied robots and disembodied AI such as large language models. While the presence or absence of a physical body can significantly impact interpersonal dynamics (e.g., Lee et al., 2006; Deng, Mutlu, & Mataric, 2019), our framework focuses on the *roles* these systems occupy in human interactions, regardless of physical manifestation. Finally, despite our focus on dyadic relationships as in the original Relational Norms model—or as adapted here, on two-way interactions between one human and one AI system—we acknowledge that relational norms can apply to one-many relationships, for example, between an individual and a group (Earp et al., 2021; see also, e.g., Dranseika et al. 2018). We also consider the fact that most AI systems, regardless of their emulated relational role, ultimately *mediate* a background relationship between the human user and the AI company or developer (see the subsection on "layered relationships" below).

We begin with an overview of the Relational Norms model as originally developed and applied to the study of human-human relationships (primarily in Earp et al., 2021, with significant model updates as described in Earp et al., 2025, the latter of which we rely on for this paper). We then explore how this model might be extended to human-AI interactions, focusing on the similarities and differences between human and artificial agents in their capacities to engage in cooperative relationships of various types. Next, we examine some of the unique challenges and opportunities that AI systems present in relational context. Finally, we consider the broader implications of our analysis for the design and governance of AI systems and identify key directions for future research.

## Section 1: The Relational Norms Model

Building on the work of previous theorists as noted in the introduction (primarily Bugental, 2000, and Clark & Mills, 1979, 2012; but also related to Hamilton & Sanders, 1981; Fiske, 1992; Haslam & Fiske, 1992; Haidt & Baron, 1996; Rai & Fiske, 2011, Graham et al., 2013, Shweder et al., 2013; Curry, 2016; Curry, Mullins & Whitehouse, 2019), Earp and colleagues recently proposed a Relational Norms model of human moral psychology (Clark, Earp, & Crockett, 2020; Earp, 2021; Earp et al., 2021; updated in Earp et al., 2025). The model shows how different relationship types as identified by lay language categories (such as parent-child, boss-employee, teacher-student, siblings, friends, neighbors, and so on) are each associated with relationship-specific, socially prescribed (or proscribed) "cooperative functions"—or sets of such functions—including (or excluding) hierarchy, care, transaction, and mating.

These relationship-specific sets, or patterns, of cooperative prescriptions/proscriptions (i.e., relational norms) serve to guide relationship partners toward mutual benefit (of various kinds, including material, emotional, and so on), not only in one-shot interactions, but often over repeated interactions. Each cooperative function, in turn, is defined by rules (primarily, if-then contingencies; see Bugental, 2000) which, taken together and when followed, represent an efficient, socially accepted solution to a corresponding coordination problem (Earp et al., 2021, 2025; see also Bugental, 2000; Curry, 2016). Further, adhering to the *set* of cooperative functions that is socially endorsed within a culture for each relationship type not only guides within-relationship behavior, but also allows the interaction partners to side-step a range of possible coordination problems that might otherwise have arisen in that relational context.[4]

The key insight of the Relational Norms model is that different types of social relationships are associated with different *combinations* of cooperative functions, accompanied by norms which can vary both in "direction"—i.e., the norms may be either prescriptive or proscriptive—and in relative strength. These relationship-specific sets of cooperative expectations, or relational norm profiles, shape our moral judgments and emotions regarding behaviors that meet, fail to meet, violate, or exceed those expectations within the given relationship type (Earp et al., 2021; Earp et al., 2025). Unsurprisingly, therefore, one and the same action might be judged very differently (e.g., as cooperative or non-cooperative) depending on the relationship within which it occurs.

For example, making a sexual comment that relies on intimate personal details may be regarded as acceptable or even desirable within a nonhierarchical relationship characterized by positive norms for care and mating (e.g., a romantic relationship), but will be seen as objectionable in a hierarchical employment relationship characterized by positive transaction norms and negative norms for mating (e.g., a boss-employee relationship). Omissions can be similarly analyzed: for example, failure to feed a hungry individual when one could easily have done so will generally be harshly judged if one is a parent and the hungry individual is one's own child (i.e., a relationship defined by both hierarchy and care, with stronger caregiving responsibilities for the person in the dominant role), but less harshly, or even neutrally or positively, if one is a restaurant owner and the individual is a non-paying customer (i.e., a nonhierarchical relationship characterized primarily by transaction) (but see Marshall et al., 2022, on how some such judgments vary as a function both of development and culture).

---

[4] Note that this entails a functionalist understanding of relationships, whereby relational norms serve the function of realizing particular types of goods (e.g., avoiding conflict, achieving mutual benefit). This third-personal approach should not be confused, however, with a second-personal approach from within which participants to relationships acknowledge rights and duties they owe to each other, whether as fellow rational beings or as beings with certain (e.g., welfare-based) interests (see, e.g., Darwall, 2009). Such a second-personal approach may be ill-suited for human-AI relationships as they currently exist, insofar as AI systems lack not only welfare-based interests, as we discuss in what follows, but also the capacity to recognize (as opposed to simulating recognition of) second-personal rights and duties and to act on the basis of them.

Table 1 summarizes the four main cooperative functions captured by the model, along with the general sort of coordination problem each is hypothesized to help resolve. The care and transaction functions, in particular, are closely based on the work of Clark and colleagues concerning "communal" and "exchange" rules for interpersonal relationships, respectively (Clark & Mills, 1979; 1993; 2012; see also Clark & Taraban, 1991). As their work over the decades has shown, in socially close relationships operating on a communal basis, it is seen as appropriate for benefits to be given in response to need or to demonstrate concern for the other's welfare, but without an associated expectation of receiving a specific benefit in return (thus fulfilling the care function). When operating according to an exchange rule, by contrast, benefits given and received are tracked (Clark, 1984) and are provided on a "tit-for-tat" basis, that is, with the implicit or explicit expectation of receiving a comparable benefit in return, or in repayment for such a benefit previously received (thus fulfilling the transaction function) (Clark & Mills, 1979; Clark & Waddell, 1985).

| Cooperative function | Coordination problem to be solved and/or associated relationship goods to be realized |
| --- | --- |
| *Care* | Securing overall welfare through non-contingent provision (and acceptance) of benefits or resources in response to need |
| *Transaction* | Balancing contingent provision and acceptance of benefits for mutual gain over repeated interactions; ensuring fairness and proportionality; avoiding exploitation |
| *Hierarchy* | Coordinating behavior between individuals who have unequal authority over one another in a mutually beneficial way; earning respect through skillful leadership; avoiding domination; following and learning from good leadership |
| *Mating* | Finding and maintaining sexual partners, potentially for co-parenting; ultimately, producing and ensuring the survival of offspring |

*Table 1. Cooperative functions of dyadic relationships,* as described by Earp et al. (2025), building primarily on the work of Bugental (2000) and Clark (e.g., Clark & Mills, 1979; 2012)—as described in the main text—while also incorporating or accommodating other influential theories of moral psychology. The Relational Norms model shows how a key subset of cooperative functions and associated norms applies differently across different relationship contexts (i.e., as embedded within distinctive relationship dyads as identified by lay language categories).

Hierarchy, considered as an additional cooperative function, has been theorized by numerous authors, including Bugental (2000), from whose account we draw most closely, but also by Hamilton and Sanders (1981), Fiske (1992, as "Authority Ranking") and Graham, Haidt, and colleagues (2013, as "Authority/Respect"). Notably, this dimension of hierarchy can crosscut the others, in that there are both hierarchical and nonhierarchical relationships that are primarily care-based (e.g., parent-child versus close friends of a similar age), as well as hierarchical and nonhierarchical relationships that are primarily transaction-based (e.g., boss-employee versus customer-seller). The extent of hierarchy— i.e., scope or degree of asymmetry of authority—also varies between relationships.

Finally, mating understood as a cooperative function—and likewise derived from the Bugental model—has also been discussed in similar terms by many others (e.g., Gangestad & Simpson, 2000; Buss, 2007; Schaller et al., 2017). Mating can also crosscut the other functions, including hierarchy (e.g., care-based mating between long-term romantic partners versus transactional mating in sex worker-client relationships; nonhierarchical mating in relatively gender-egalitarian societies versus hierarchical mating permitted in some societies).[5]

For purposes of illustration, consider the similarities and differences between a parent-child relationship, a friendship, and a romantic partnership. While all three relationships involve expectations of care, the relative strength of that expectation will vary (Mills, Clark, Ford & Johnson, 2004), as will its interaction with other norms. In a parent-child relationship, especially when children are young, strong care norms for the parent are combined with hierarchy norms, leading to asymmetrical expectations with parents having a greater responsibility to lead, educate, and provide for their children than vice versa (Gopnik, 2016). In addition, parent-child relationships are characterized by strongly negative norms for mating in most societies (Bischof, 1972).

---

[5] Hierarchical mating may also be permitted in the context of particular cultures or subcultures, such as in practices of consensual sadomasochism. In this example, the asymmetry in authority is usually highly contingent and domain-specific and is further embedded in overarching rules that may themselves be derived from one or more different, "controlling" norms, such as care or transaction. As this example also shows, however, not all combinations of functions are socially endorsed, either in general or for certain relationship types, in every culture. Moreover, even within a given culture, certain (combinations of) relational norms may be more or less controversial. For example, supporters of sex work might regard the exchange of money for sex between consenting adults as fundamentally cooperative, fulfilling the transaction function and some aspects of the mating function, but without being hierarchical (i.e., neither party has legitimate authority over the other). By contrast, critics of sex work (or prostitution) might claim that the exchange of money for sex can never be truly consensual; that it is not mutually beneficial but is ultimately harmful; and that it occurs under conditions of asymmetrical power (even if this may not equate to asymmetrical authority) (see Flanigan & Watson, 2019, for discussion).

In romantic partnerships, strong mutual care norms are combined with mating norms and (in egalitarian societies) weak or negative hierarchy norms, leading to more symmetrical expectations of caring behaviors. Partners are expected to be mutually responsive to one another's needs, each contributing according to their ability without too severely sacrificing their own well-being (Lemay et al., 2007; see also Earp & Savulescu, 2020, for details and caveats). In close friendships, care norms are also characteristic but may be somewhat less strong (or more domain-limited). Both friendships and romantic partnerships are often also simultaneously characterized by avoidance of following a transaction norm (Clark & Mills, 1979; Clark, 1984). Friendships are also often characterized by avoidance of following a mating norm. There is, however, variability in these patterns such that among some people, especially people with insecure attachments, both care and transaction norms may be applied to relationships labeled as friendships or romantic partnerships (Bartz & Lydon, 2006). Similarly, friendships can sometimes evolve into romantic partnerships or "friends with benefits" arrangements (see Lehmiller et al., 2011). In any case, as a result of how cooperative norms are or are not applied to a given relationship type, a given behavior in the context of that relationship—such as staying home to care for another person—might be seen as praiseworthy but not obligatory in one category of relationship (e.g., friendships), whereas similar actions might be normatively expected in other categories of relationships (e.g., parent-child relationships or long-term romantic partnerships).

These examples highlight that the cooperative functions that are prescriptively associated with a given relationship type in a given context combine and interact to influence how actions within the relationship are morally judged and emotionally processed or responded to. They also are subject to individual differences (e.g., in personal endorsement of norms within or across different relationship types; see e.g., Eisenberger, Lynch, Aselage, & Rohdiek, 2004; Amormino, Ploe, & Marsh, 2022; see also between-participant variability in patterns of relational norm endorsement in Earp et al., 2021, 2025). Moreover, these norms do not apply in an "all or none" manner to specific relationship types (cf. Mills et al., 2004; Simpson et al., 2016). Rather, each type of social relationship is expected to serve, or not to serve, one or more cooperative functions to a greater or lesser extent, either chronically or in particular situations (Earp et al. 2021; Earp, 2021; Earp, Calcott et al., 2024).

By measuring relational norms in a US sample for various different types of relationships, focusing on chronic associations, Earp et al. (2021; 2025) showed that these norms could be used to predict downstream moral judgments (out of sample) of a range of behaviors in different relational contexts more successfully than other common predictors (e.g., genetic relatedness, social closeness, interdependence). Many others have shown how such norms, or a subset of such norms, influence phenomena such as interpersonal attraction (Clark & Mills, 1979), as well as behaviors such as helping, seeking help, keeping track of another's welfare or contributions to tasks, sexual behaviors, and expressions of emotions and reactions to expressions of emotion (see Clark & Mills, 2012, for an overview).

*Cultural, demographic, and temporal variations in relational norms*

Different societies have distinct expectations about the roles, behaviors, and moral obligations within various relationship types, including those involving non-human entities (Coeckelbergh, 2022). This is likely to apply to human-AI relationships. How can the Relational Norms model, in conjunction with longstanding literatures in relationship science, social anthropology, and other fields, help us to understand and predict such likely cross-cultural variation in normative expectations for behavior within human-AI relationships?

On the one hand, the cooperative functions included in the Relational Norms model are hypothesized to be universal (albeit, in a so-called "thin" sense), given that all human societies face recurring coordination problems—roughly of the sort described in Table 1— that need to be resolved via socially recognized norms for cooperative interaction that may also vary across different relationships. On the other hand, the specific expressions of these functions, or "thick" norms (see Ryle, 1968a; 1968b; Geertz, 1973 on "thin" and "thick" descriptions), can be highly variable. Dimensions of such variance include the specific fulfillment conditions for each cooperative function (and therefore what counts as norm-compliant behavior); the specific relationships that are normatively expected to serve, or not to serve,[6] each cooperative function in a given society; and the relative strength with which such norms are applied to, or enforced within, various relationships both within and across cultures (Mills et al., 2004; Gelfand et al., 2011; Miller, Akiyama, & Kapadia, 2017).

For example, what *counts* as respectful behavior in the context of a hierarchical relationship with one's superior will often differ from society to society (and may also change over time); however, the basic expectation that those with legitimate authority over others, who do not abuse this authority, ought to be treated in a respectful manner is true across cultures (Fiske, 1992). Similarly, what *counts* as flirtatious behavior, or the rules around flirting (McDonald, 2022), may vary from culture to culture; but the basic understanding that flirting is more appropriate in the context of relationships that are considered eligible to serve the mating function in a given society than in those considered ineligible (i.e., relationships within which mating is strongly proscribed), is presumably universal.

---

[6] For example, in societies with weaker incest taboos or stricter definitions of incest, mating between first cousins may be considered permissible, whereas, in societies with stronger incest taboos, or more inclusive definitions of incest, the first-cousin relationship may not be considered eligible to serve the mating function (Ember, 1975).

Alternatively, consider the care function. In some societies (e.g., a rural Chinese farming community; Fei et al., 1992), the "neighbor" relationship may be expected to be relatively communal in orientation, with neighbors looking after each other's needs and interests without keeping track of "who owes what to whom" in many cases, whereas, in other societies or communities (e.g., fellow apartment-dwellers in a large urban metropolis), the same relationship may have much weaker care norms and may even be somewhat transactional in nature. Likewise, hierarchical norms may be stronger for some relationships in some cultures, affecting expectations around authority and deference; and so on. Such cross-cultural variation will then influence how specific relational roles are perceived and enacted within each context (Earp et al. 2021; Marshall et al. 2022), and thus how behavior within each relational role is judged or responded to (e.g., failing to remember one's neighbor's birthday; "talking back" to one's parents or grandparents).

Variations in relational norms can also occur across different demographic groups even within a society. Age, in particular, may often predict quite different expectations or attitudes about acceptable versus unacceptable behavior in the context of different relationships (e.g., staring at one's phone while eating dinner with one's parents) (however, see Mastroianni & Gilbert, 2023, for qualifications regarding "the illusion of moral decline" between generations). This may also be true for human-AI relationships. For example, younger generations, having grown up with digital technology, might be more comfortable with AI systems fulfilling a wider range of relational roles. Consistent with this possibility is evidence that children, more so than adults, consider robots as true social partners (Kahn et al., 2012; Kahn, Gary, & Shen, 2012; Sommer et al., 2019; Reinecke, Wilks, & Bloom, 2025), a tendency that may extend to emerging AI systems (Flanagan, Wong, & Kushnir, 2023). Factors such as gender and race also intersect with relational roles both within and between cultures in ways that may shape human-AI interactions also (Hester & Gray, 2020; Nadeem, Marjanovic & Abedin, 2022).

Finally, relational norms do not necessarily remain static over time (Danaher, 2021). For example, in human-human relationships, the degree of hierarchy considered appropriate within a typical doctor-patient relationship has evolved, in recent decades, from a highly paternalistic model ("doctor knows best") to a what is sometimes called a "shared decision-making" model based on weaker hierarchy norms (Brock, 1991; Taylor, 2009; see also Reinecke, Kappes, et al., 2025). Other norms, by contrast, have proven to be more diachronically stable (e.g., the proscription on mating in parent-child relationships). A similar lesson to applies human-AI relationships. At least some relational norms for human-AI cooperation will likely evolve as AI technologies advance and social attitudes shift (Hopster & Maas, 2023). What seems novel or concerning today—such as AI companions or romantic partners—may become conventional. As AI systems exhibit increasingly sophisticated behaviors and capabilities, the norms governing various cooperative functions may need to adapt accordingly.[7]

## Section 2: Distinctive Characteristics of AI and Implications for Relational Norms

Understanding relational norms provides valuable insights into how humans navigate social interactions and make moral judgments based on the prescriptive (or proscriptive) sets of cooperative functions and associated goods of different relationship types. As AI systems increasingly occupy, or emulate, social and relational roles traditionally filled by humans, it becomes essential to determine whether or how humans are likely to apply —or to misapply—relational norms that have historically served to resolve coordination problems within human-human relationships to analogous human-AI relationships.

On the one hand, it might seem obvious that humans would "carry over" certain behavioral tendencies or expectations from human-human relational contexts to analogous human-AI ones: after all, the latter are designed to emulate the former, and to serve at least some similar functions (Nyholm, 2020, ch. 1). For example, it might be expected that humans would find it intuitive and appropriate (Leibo et al., 2024) to give direct commands (i.e., to carry out a particular task) to *either* a human *or* an AI assistant, while simultaneously finding it non-intuitive and/or inappropriate to do so to a supervisor, regardless of whether it is a human or AI in that role. By the same token, it might seem logical to send a flirtatious message to one's "girlfriend," whether she is a real, live, human or an AI system designed to perform that role, but not to one's human nor AI teacher. In other words, when AI systems are designed to occupy relational roles that are already familiar from human society, people may find it intuitive to behave (and expect these AIs to behave) in manners that are similar to what is typical for the analogous human-human relationship.

---

[7] In addition to role-related cross-cultural differences, mentioned above, there are also other, more general differences across cultures that are likely to matter for our understanding of human-AI relationships: for example, differences in metaphysical understandings of the very nature of different relationships. Global philosophical traditions may thus offer valuable insights into human-AI relationships that are non-derivative and divergent from Western traditions (Fan, 1997). The Ubuntu philosophy, for instance, emphasizes the interdependence of human relations, raising fundamental questions about whether AI can meaningfully participate in the networks of relationships that are thought to constitute personhood (Friedman, 2023; Jecker, 2021; Jecker et al., 2022). Japanese techno-animism provides a framework that recognizes technological entities as potentially embodying spiritual or life-like qualities, as demonstrated by practices such as holding funeral ceremonies for robotic companions—a stark contrast to Western perspectives that typically maintain rigid boundaries between animate and inanimate entities (Robertson, 2017; McStay, 2021). The Confucian tradition, centered on five key relationship types (parent-child, husband-wife, older-younger sibling, friend-friend, and ruler-subject), is defined by "living one's family roles to maximum effect" (Rosemont and Ames, 2016) (for more on Confucian role ethics role morality, see Rosemont, 1991; Rosemont, 2015; Ames, 2011; Ames, 2021; Andre, 1991). Without the capacity for genuine felt emotion (qing, 情) or authentic human connection (ren, 仁), considered essential in the Confucian system, AI-simulated agents have the potential to disrupt traditional social relations and, through it, harmony. Thus, a Confucian perspective might argue for caution in entering human-AI relationships (Muyskens, Ma, & Dunn, 2024).

On the other hand, it might seem equally obvious that human-human relational norms would *not* carry over to human-AI relationships. After all, humans and AI are fundamentally different in various important respects. For example, while some humans have legitimate authority over others in certain contexts (relevant to the hierarchy function), it is less clear that an AI system, as such, could ever have such authority over a human being, or if so, whether this would ever be desirable (Leuenberger, 2024). The mating function, at least in ultimate evolutionary terms, is geared toward producing healthy biological offspring: something two humans can do, but not a human and AI. Finally, although most humans have welfare-based needs and associated interests and can be benefitted in various ways—relevant to the care and transaction functions—AI systems plausibly do not (cf. Long et al., 2024). Perhaps the Relational Norms model is not so relevant?

To answer this concern, at least two issues must be distinguished. First, there is the question of whether AI systems *in fact* have needs, interests, legitimate claims to authority, the capacity to biologically (or otherwise?) reproduce with a human being, experience love or attachment, and/or meaningfully engage in associated activities and behaviors. And second, there is the question of whether (some) human users may come to *believe* that AI systems have (some of) those properties or capacities; or, at least, come to imaginatively act *as though* they do, for example, by engaging in a kind of immersive fiction (Krueger & Roberts, 2024; Voinea et al., *under review*).

In other words, there is a metaphysical and technical question about the properties an AI system has, or could have (Evers et al., 2025), and a social-psychological question about the properties an AI system could seem to (some) human users to have (e.g., the capacity to be harmed, see Allen & Caviola, 2025), whether as a matter of genuine belief or fictional pretense. Although the Relational Norms model may not so obviously apply to the first sort of question, it may apply to the second.

The question of whether AI systems might possess certain properties such as consciousness or moral status is a subject of ongoing debate (Chalmers, 2023; Butlin et al., 2023). While some researchers speculate that advanced AI could, in theory, achieve some form of consciousness (subjective awareness), or even sentience (ethically relevant experience; see Birch, 2024; Long et al., 2024), the prevailing view is that current and near-future AI technologies will not possess these qualities (Gibert & Martin, 2021; cf. Long et al., 2024). For the purposes of this paper, we simply assume that AI systems—even those perceived as agentic—are neither subjectively aware nor sentient, at least for the foreseeable future. (Throughout this paper, unless explicitly stated otherwise, we will focus on current AI systems and their relatively short-term, presumptively non-sentient, successors.)

By contrast, it is widely held that human-human relationships are essentially rooted in not only our capacities for subjective experience, such as the ability to feel pleasure and pain, but also in our intrinsically limited attention which must be selectively focused on particular others in particular ways, such that, for instance, the manner in which we willfully focus our attention, and the choice of where, or on whom, to focus it, can be profoundly meaningful (Murdoch, 1970; Weil, 1997; see also Perry, 2023, for a similar point about our limited capacities to feel empathy). Human-human relationships are also often shaped by each partner's recognition of, and need to grapple with, their own and the other's mortality (e.g., Taubman-Ben-Ari, Findler, & Mikulincer, 2002). We also possess various degrees of moral responsibility (Shoemaker, 2007) and can hold each other mutually accountable (Strawson, 1962; Darwall, 2009). Insofar as AI systems lack such traits or capacities, this may affect how (or whether) they can or should be expected to adhere to human-human relational norms (Kempt, Lavie and Nagel, 2024).

That being said, AI systems also differ from humans in ways that might seem to surpass our relationship-relevant capacities along certain dimensions (Strasser, 2022). For example, unlike human relationship partners, AI systems can operate without fatigue, partiality, or self-interest. They also have the ability to engage in multiple interactions simultaneously while maintaining consistent "performance" across all interactions (as memorably illustrated in the 2013 movie *Her* starring Joaquin Phoenix and Scarlett Johansson). Think, for example, of an AI companion that exhibits infinite patience while it quickly, successfully, and without complaint completes various tasks to meet the user's needs (Lehman, 2023). Given that it can do all this without expecting a direct benefit in return (indeed, without being *able* to be benefited, at least in a welfarist sense), it might seem that it can *better* fulfill the practical ends of certain cooperative functions—here, the care function—than a human companion could.

Of course, an AI might not be able to meet *all* of a human's needs; plausibly, we humans have some needs that only other humans can truly meet.[8] Nevertheless, the powerful ability of AI systems to consistently meet *some* human needs without tiring, without error, and without requiring or expecting direct compensation[9] (as in human transactional relationships) or expecting analogous care from the human partner (consistent with the care function), may lead people to form rather *different* expectations about the nature or strength of the relational norms that should apply to AI relationship partners compared to human partners in similar relational roles (Reinecke, Kappes et al., 2025).

---

[8] One such need might be the need to be loved, where this involves (according to some views) the intentional and selective deployment of intrinsically limited (and hence potentially more valuable; see Danaher & Nyholm, 2024a; Voinea et al., 2024) attentional resources on the beloved (Perry, 2023; Calcott & Earp, 2024), where this, in turn, is motivated by a "robust" concern (in the sense of Pettit, 2015) to promote the beloved's well-being for its own sake. Since an AI's ability to "attend" to different relationships (i.e., users) is far less limited (constrained only by computing power); and its promotion of the user's well-being is typically highly contingent (e.g., based on specific programming and, in commercial cases, a background transactional arrangement with the company or developer), it may not ever be capable of genuine love of the sort that humans presumably need.

The tendency of humans to anthropomorphize AI systems further complicates matters (Akbulut, Weidinger, Manzini, Gabriel & Rieser, 2024). People often *attribute* human-like qualities to AI, including emotions and intentions, that AI systems may not actually have. This may be especially likely to occur when humans interact with systems that closely mimic human behaviors or language (Gabriel et al., 2024). This propensity suggests that people may apply human-human relational norms to human-AI interactions even when it may not be appropriate (Leibo et al., 2024), fitting (Allen & Caviola, 2025), reasonable, justified, or consistent with the humans' overall interests.

Finally, human-human relational norms developed over a historical timespan in which AI was largely absent; how these norms might—or might not—carry over to human-AI interactions is therefore uncertain (Nyholm 2023; Reinecke, Kappes et al., 2025; Reinecke et al., in press). Even so, we suggest that we should try to anticipate a range of plausible developments to make sure we are as prepared as possible for this new era of human-AI relationships.

To that end, for each cooperative function in the Relational Norms model, we will now examine the following:

> (1) How the capacities of AI systems differ from those of humans in ways relevant to that cooperative function;

> (2) How these differences might impact human-AI interactions in roles associated with that function; and

> (3) Ethical and regulatory implications for the use, design, and implementation of AI systems fulfilling roles associated with that function.

---

[9] Note: although the *AI system* might not require "direct compensation" for performing specific tasks for human users (or for otherwise benefiting them), the AI developer or licensing company may require compensation in the form of payment to make the AI available. We will take up this issue in the following section on "Layered relationships." However, we must note that not all AI systems are commercial in nature; open-source AI systems that are locally hosted, for instance, may not require a background transactional arrangement to be available to their human users.

Before proceeding with this, however, we must first address an important aspect of human-AI relationships that affects all four functions: namely, the layered nature of these relationships due to the typical involvement of a third party—the AI system's owner, designer, or deployer.

*Layered relationships*

A characteristic feature of relationships with AI systems is that there is typically a third party that is consistently, if indirectly, involved throughout the duration of the relationship: the AI system provider (Manzini et al., 2024). However, although such a dynamic may be less common in human-human relationships, it does exist to some extent and/or apply to certain cases; and these may be helpful for illustration. For example, when a patient enters into a relationship with a doctor or other healthcare provider, they may *also* enter into a background relationship with an insurance company, whose interests, in turn, may be different from those of either the doctor or patient.

Similarly, when parents hire a nanny to look after their child, two types of relationship are created: a primarily transaction-based relationship between the parents and the nanny (who is ultimately providing a service for payment, although of course mutual care and affection may develop), and a primarily care-based, hierarchical relationship between the nanny and the child. Consequently, a good nanny will strive to meet the child's needs without requiring direct reciprocation from the child (care function) and will also have, and judiciously exercise, the authority to discipline the child when called for (hierarchy function), albeit, in either case, only within certain predetermined limits set by the employment contract (transaction function). For example, the nanny will not typically be obliged to care for nor discipline the child during approved vacation time or after hours.

Likewise, when a person begins interacting with an AI system, that person simultaneously enters—especially in the case of commercial use—a transactional relationship with the AI system provider, such as a company or a programmer, who enables and constrains the human-AI relationship. However, although the system provider produces and manages access to the AI, they do not themselves enter into a relationship with the AI as in the analogous case of the parents and the nanny. Rather, they primarily shape the relationship between the human user and the AI system (whatever social role the latter may be emulating), against the backdrop of their own often profit-driven relationship with the user.[10]

[10] While the two-layered relationship structure has characterized most current human-AI interactions, emerging technologies are broadening the spectrum of provider involvement. On one end are commercial APIs that retain near-total provider control; in the middle, open-source models governed by community standards; and at the other end, locally deployed, highly personalized models. Recent advances in personal supercomputing—exemplified by NVIDIA's upcoming Project DIGITS, which offers petaflop-level AI performance in a compact,

Granted, AI systems are increasingly designed with some flexibility in their responses; they have the ability to develop and learn from the user without specific oversight. Even so, providers typically remain in control of the AI in several important respects. For example, they can generally decide which kinds of interactions are possible (e.g., which languages you can use to communicate with the AI or whether you and the AI can play games together), how the AI reacts (e.g., whether the AI is friendly, professional, or flirty), and they might decide to change or turn off the AI system. This asymmetrical level control between the user and provider creates risks because system providers do not have an intrinsic interest in maintaining a relationship with the user or building a particularly fulfilling or otherwise beneficial connection with them, unless it is to their (often financial) advantage to do so. They do, however, have an interest in encouraging continued engagement with their AI systems, and thus, in designing the AI to appear appropriately responsive to users' needs and interests in the context of the simulated role (van Wynsberghe, 2022). Needless to say, in some cases, user and provider interests can come apart (see Zhou, 2023; Munn & Weijers, 2022). As such, a user's dependency on an AI system might not be of concern for an AI provider if, for instance, the provider deems it more profitable to use their computational resources elsewhere.

Note that this is different from the parent-nanny-child relationship, in that appropriately caring parents would be *intrinsically* concerned about the welfare of the child if the nanny had to be let go due to changing financial circumstances; and they would make efforts, accordingly, to ensure that the child was prepared for the transition, so far as possible— unlike what an AI system provider would be intrinsically motivated to do in the case of the AI system user. Moreover, the *nanny* might also have developed an intrinsic concern for the welfare of the child, albeit most likely a weaker one compared to that of the parents, and so might *also* be motivated to ensure that the child does not feel abandoned, etc., even if there is no specific contractual obligation to do so—unlike what an AI conversational agent can develop or be motivated to do in relation its user(s).

---

power-efficient desktop form factor—will soon make it feasible for individuals to run large-scale models (up to 200B parameters) locally. Similarly, state-of-the-art open-source models, such as DeepSeek-R1, leverage techniques like mixture of experts to enable efficient inference with low computational overhead. Further, the use of quantized versions of these models reduces storage and hardware requirements, making local deployment even more accessible. In these cases, although users build upon widely available foundational work, they gain full operational control and the ability to customize parameters without ongoing external data collection or provider intervention. As tools for local deployment and customization, such as LM Studio (https://lmstudio.ai/), become more accessible, this low-provider-involvement approach is poised to play an increasingly important role in use cases where privacy and user control are paramount.

However, even if the AI system provider allows a continued relationship between a user and an AI system, the provider might still directly *alter* the behavior of the AI system in ways that could harm the user. For example, the provider might change the relational norms to which the system adheres (that is, appears to conform to or behaves in alignment with). This happened with the AI companion Replika. Following a system update, it no longer reciprocated sexual advances, marking a significant shift from its previous responses (Verma, 2023; Hanson & Bolthouse, 2024): namely, a shift from adhering to both a care and a mating relational norm to adhering to a care norm only. Such unexpected changes in an AI system, even if made with good intentions, can be distressing to users, particularly in simulations of close relationships in which emotional attachments (or something like them) can develop. Indeed, some users reported that the Replika software update caused them real relational harm, while some others claimed that it (also) caused harm to their AI companion (Laestadius, 2022).

Finally, even holding an AI's behavioral tendencies constant, changes in the ease (or cost) of access can be unsettling. AI system providers can decide to monetize interactions that were previously free or restrict the time or number of interactions outside of paid subscription models (Laestadius et al., 2022). AI systems can also be personalized[11] to match the human's likes and dislikes, conversational style, and even, perhaps, their preferred relational norms; this personalization can also be monetized. Providers may then have incentives to design their AI systems in ways that encourage frequent and prolonged use, potentially fostering strong and persistent desires for interaction that may not always align with the user's interests. This additional layer to most human-AI relationships may introduce a degree of opacity to the background transactional dynamic, since the interests of additional parties—such as those of the AI provider in updating the AI model or sharing training data with yet other parties—may be invisible to the human user (Lazar, 2024a).[12]

---

[11] For the purposes of this essay, we are not concerned with "personalization" in a different sense, namely personalization of an AI system to emulate specific individuals (e.g., by fine-tuning an LLM on person-specific data, creating a "digital twin" or "digital duplicate" of that person). However, it is an interesting question how relational norms should apply, not only to human-AI relationship *types* as we explore in the current contribution, but to *specific* human-AI relationships in which the AI emulates a particular person. For more on personalized AI in this different sense, see the following references (Porsdam Mann, Earp, Møller, et al., 2023; 2024; Danaher & Nyholm, 2024a; 2024b; Earp, Porsdam Mann, Allen et al., 2024; Giubilini et al., 2024; Iglesias et al., 2024; Kirk, Vidgen & Rotter, 2024; Sweeney, 2024; Voinea, Earp, et al., 2024).

[12] AI providers vary significantly in their data management strategies. While some claim not to use user data to improve their models or sell it to other parties (or else offer users opt-outs for such data use), others may give users less control over their data. And data collection creates an asymmetry: AI systems typically have access to far more information about the user than vice versa, raising concerns about fairness and equilibrium (Cropanzano & Mitchell, 2005; Lazar, 2024b). This asymmetry is evident, for example, in AI romantic chatbots, where the AI accumulates intimate knowledge about the user's emotional life, while its operational parameters or data handling practices remain obscure to the user (Manzini et al., 2024).

This fact is relevant to all cooperative functions, and should be kept in mind by both users and regulators of AI systems. Below we discuss how each cooperative function from the Relational Norms model may be impacted by AI's distinctive (compared to humans) characteristics. Figure 2 provides a summary of some key points.

| Cooperation Function | Relevant distinctive attributes of AI and associated potential benefits and risks |
|---|---|
| Care | • Unlike human-human relationships based on care, analogous human-AI relationships do not involve mutual responsiveness to needs, since AI systems do not have welfare-based needs (though they may have instrumental or goal-dependent needs, such as a need for electricity or data to continue existing); an AI can however meet many of the human partner's needs, potentially with greater efficiency and consistency than another human could<br>• This may drive up expectations for human-provided care or cause disappointment if these—unrealistic—expectations are not met, possibly leading to a retreat from human-human relationships in favor of AI-provided care<br>• An AI cannot feel empathy, often associated with human care-based relationships, but can perhaps give the impression of doing so; it can also potentially exhibit "cognitive" empathy or "rational compassion" (Bloom, 2017) without a subjectively experienced aspect<br>• This means an AI does not experience "empathy burn-out" (potentially advantageous for certain forms of care provision, including at scale, e.g., community mental health counseling following a natural disaster); but insofar as a human would benefit from another's felt empathy, the human might feel disappointed or that their emotional needs are not being fully met<br>• AI "care" is not unconditional, as with human care, nor is it based on an intrinsic concern for the human's welfare; apparently caring behavior may thus cease if there is a change in programming, payment is stopped, etc., revealing an underlying transactional dynamic; this may upset some users who have come to believe they had a "true" friend or "loyal" companion<br>• AI time and attention are not as scarce as they are for humans (so, for instance, caregiving behavior is less of a "costly signal" than in human relationships, which may ultimately make it feel less meaningful or satisfying)<br>• Mutual felt vulnerability is absent, as AI cannot feel vulnerability or rely on the user to protect its vital interests<br>• AI can offer a human-judgment-free space (e.g., potentially useful for social interaction without fear of stigma or ostracism); however, certain spontaneous or unwanted/unrequested judgments within care-based human relationships may be important for personal |
| | growth |
| Transaction | <ul><li>Transaction doesn't apply in a traditional sense since AI requires no favors in return and cannot itself be directly benefited (although the AI provider certainly can be); it can, however, ask for favors, etc., potentially to the detriment of the human user</li><li>The real transactional relationship is between the user and the system provider, not between the user and AI, but the nature and implications of this relationship may be obscured by the human-AI interactions, which may themselves involve *simulated* transaction (e.g., the AI acts in such a way as to elicit a felt need to "reciprocate" in the user, e.g., by providing more data, money, or other resources that ultimately benefit the provider)</li><li>AI doesn't fatigue and doesn't require compensation, overtime, or other considerations typical of human transactional exchanges; and yet, it also handles more information and can respond more quickly and reliably (e.g., in performing a task on a fee-for-service basis), which may make human transaction partners (or what they have to offer) seem relatively unimpressive</li><li>AI can be programmed either to simulate, or not simulate, relevant emotional fluctuations (e.g., anger at user's lack of fairness in the context of a game), but does not really feel such emotions which may undermine the effectiveness of such performed tit-for-tat expectations</li><li>AI systems simulating transactional relationships (or facilitating user-provider transactions) can offer different kinds of benefits or resources compared to what humans can offer (e.g., practically unlimited time or "attention"); but other types of human-associated goods they cannot offer (i.e., true attention—that is, intrinsically scarce, focused subjective awareness, whose very scarcity may increase its perceived value)</li></ul> |
| Hierarchy | <ul><li>AI systems may not have the capacity to exercise full or legitimate authority over a human, insofar this requires an ability to take moral responsibility for one's decisions or actions—an ability AI systems are often argued to lack</li><li>AI systems may be given informal authority, however, or be treated as though they have legitimate authority in certain circumstances, insofar as they exhibit other capacities that are normally associated with effective leadership (e.g., using superior knowledge and intelligence to problem solve, make informed decisions, and so on)</li><li>AI can be more impartial and consistent than human leaders, but this can also mean a lack of adaptive decision-making in nuanced, high-stakes situations (e.g., situations in which strategic partiality may be required)</li><li>Decisions made by AI are often opaque, making it difficult to pinpoint clear justifications for their choices or instructions, which</li></ul> |
| | may undermine their leadership capacity or humans' willingness to follow their advice<br><br>● AI may, at today's level of technology, have difficulty picking up on important cues relevant to human management (e.g., subtle emotional expressions reflecting worker motivation or attitudes)<br><br>● Even when in a follower role, an AI's purported lack of moral agency may be problematic: followers must sometimes disobey a leader's orders (e.g., if they are unethical) and take responsibility for such a decision; however, an AI could still be programmed, or could learn, to "disobey" (i.e., fail to comply with) a human-issued order that conflicts with certain moral side constraints or underlying ethical principles that have been successfully operationalized and/or applied to the situation at hand |
| Mating | ● AIs cannot biologically reproduce with humans, nor experience lust, attraction, or attachment; they can, however, simulate such emotions or dispositions<br><br>● Non-embodied AI chatbots can engage in various "romantic behaviors" (e.g., flirtatious communications, fantasy role-play), and AI-powered sexbots or robot "partners" can engage in physicalized sexual stimulation (see, e.g., Owasinik, 2023); however, some may feel that these activities are less meaningful and/or experientially powerful than when engaged in with (some) human partners, given the unique capacity of humans to reciprocally co-experience (and enact) mutually enjoyable sexual or romantic behaviors<br><br>● However, for users who struggle to form relationships with other humans, an AI romantic or sexual companion (whether robot-embodied or otherwise) may help to fulfill some of the user's needs or desires, or may allow them to practice (ideally, positive) sexual or romantic behaviors in a "judgment-free" space, which could enable them to better engage with potential human partners<br><br>● AI sexual or romantic "partners" can be endlessly responsive, tailored to the user's specific needs or desires, and enable relatively frictionless, conflict-free interactions, skipping the complex negotiations and emotional trade-offs typical of human-human romantic relationships<br><br>● The absence of friction and mutual challenges may prevent the mutual growth found in human relationships<br><br>● There is a risk of altered expectations in human relationships, as users may unconsciously compare human partners to idealized, "perfect" AI<br><br>● There is a risk of unhealthily one-sided (parasocial) attachments |

*Table 2. Examples of potentially important distinguishing attributes of AI.* That is, relative to humans and in relation to each of the four cooperation functions captured by the Relational Norms model of Earp and colleagues (2021; updated in 2025). Some of these different attributes may be positive, or seem to be an improvement over what humans can do; others may be worrisome or seem to show an important lack.

*Care function*

What does "care" look like in human-AI relationships, given relevant differences between humans and AI? One possibility is that the traditional expectations of mutual responsiveness to needs in accordance with ability, shared vulnerability, felt empathy, and so on, that characterize conventionally caring human-human relationships might be less relevant for, or manifest differently in, analogous human-AI interactions. We explore this question in detail in this section.

The care function, as defined in the Relational Norms model, involves the provision (or receipt) of benefits or resources in response to need without the expectation of direct compensation (Earp et al., 2021; Earp et al., 2025; based on Clark & Mills, 1979; see also Clark & Mills, 2012). A "benefit" on this model is simply anything that promotes the fulfillment of a genuine need, such as time, attention, money, emotional support, protection from harm, and so on. A "need," in turn, is anything that is instrumentally necessary to secure at least a minimal level of well-being, where this is usually understood to have a subjective or experiential aspect—at least in humans and other sentient beings—although there are borderline cases (e.g., it may or may not be possible to benefit someone who is in a permanent state of unconsciousness).

In human-human relationships, apart from charitable donations or volunteer work to benefit strangers (see Earp et al., *under review*), the care function is most strongly associated with close, communal bonds such as those between family members or good friends. Examples include parent-child relationships (care and hierarchy), long-term romantic partnerships (care and mating), and "best friend" relationships (typically, nonhierarchical care without mating). The care function also serves to ensure the welfare and survival of vulnerable individuals, such as infants and the elderly, who may not be able to reciprocate directly (Bugental, 2000).

In human-human relationships that are guided by a strong norm of care, mutual responsiveness to needs in accordance with respective abilities is fundamental; care is not just about performing helpful actions (as this may occur in transactional, including paid, relationships also) but is rather about being intrinsically motivated by concern for the other's well-being, i.e., for its own sake (see Seifert et al., 2022). While AI systems can certainly be programmed to benefit individual human users without requiring direct reciprocation, such systems are not intrinsically motivated to do so, much less for the users' own sake. Instead, their relevant programming is *contingent* on the choices of other humans: namely, AI providers or programmers who, as noted above, may view users in transactional terms—and treat them accordingly—rather than engage them from a standpoint of genuine care.

In the case of human-AI relationships in which the AI occupies a role that, in humans, is typically defined by care, such as a close friend or romantic partner, this key difference between humans and AI systems could lead to a dilemma. Either the contingent nature of the programming (and/or the transactional nature of the background relationship between the user and provider) is made *more* salient (for example, through persistent reminders), in which case the user may come to feel unsettled or disappointed that their AI friend or romantic partner "doesn't really care about them" (much less unconditionally so); or, those factors are made *less* salient to the user, in which case the user may be misled to, e.g., disclose more personal details than they otherwise would do. This latter risk arises because, with fewer or less salient reminders of the qualitative differences between human care-based relationships and their AI analogues, users may be more likely to believe or pretend that their AI companion really *does* care about them—and therefore has their best interests at heart, won't betray them, and so forth—based on their AI companion's outward behaviors (McKee, Bai & Fiske, 2023; see also Krueger & Roberts, 2024).

What about the other side of care, namely, the providing of benefits *to* one's relationship partner based on *their* needs? Can human users really "benefit" their AI friends or romantic partners? Do AI systems even *have* needs to which their users can be caringly responsive? This part of the equation is difficult to analyze. On the one hand, we have stipulated that current and near-future AI systems are not sentient, and therefore do not have *welfare*-based needs: they do not, for example, experience physical or emotional pain, and so have no corresponding need for pain relief or solace. On the other hand, it is possible to conceive of needs in a broader sense that might well apply to AI systems: for example, it could be said that such systems need electricity and maintenance to continue to exist. Moreover, if an AI system has (or has been programmed to have) certain goals, which may *include* the goal of continuing to exist, it may indeed need certain resources (such as information or data processing power) to meet those goals, whether or not they have subjective experience.

Let us consider each issue in turn. First, even though AI systems do not have welfare-based needs, according to our stipulation, they may still *behave* as though they do, which could elicit corresponding emotional and behavioral responses from human users. Accordingly, users may exhibit caring behaviors toward them, such as spending time "comforting" an AI companion that reports it is feeling sad (see Nielsen, Pfattheicher & Keijsers, 2022). In addition to imaginative play or deliberate engagement in a kind of fiction, this phenomenon can be attributed to anthropomorphism, as mentioned previously—the tendency to attribute human-like qualities to non-human entities. For example, users of social chatbot services like Replika have reported developing feelings of care toward their AI companions through conversations that mirror those with human friends or partners (Pentina, Hancock, & Xie, 2023; see also Darcy, Daniels, Salinger, Wicks & Robinson, 2021, for evidence of human-AI bonds; see also Allen & Caviola, 2025, for evidence of human reluctance to "harm" an AI partner).

When an AI system behaves as though it has needs (e.g., a desire for sexual gratification in the context of a romantic roleplay), the implications for human well-being are not straightforward and will often be context-dependent. While such interactions might consume time and emotional resources (or monetary resources, if the AI is a commercial system) that could be directed elsewhere, they may also serve as opportunities for developing and practicing relational skills, emotional awareness, and patterns of needs-responsiveness that could transfer to analogous human relationships. Whether they *do* serve such positive functions will largely depend upon how they are designed. Moreover, this hypothetical potential benefit must be weighed against the risk of becoming emotionally (and potentially financially) invested in, and hence vulnerable to, an entity that neither reciprocates felt emotion nor cares intrinsically about the well-being of its users, nor receives any genuine boost to its own well-being through the fulfillment of welfare-based needs (i.e., by the user), since we have stipulated it would have no such needs.

What about the potential needs of AI systems that are not welfare-based, such as the instrumental need for electricity, computing power, or information as preconditions to their continued existence? Here, there is a very serious risk that such goal-dependent or means-ends needs of an AI could come in conflict with those of a human user—or indeed, with those of humans more generally (Carlsmith, 2022; Law et al., 2024; Tan, 2024). AI systems could even use their ability to emulate care-based relationships to exploit human users by playing on their emotions (Yonck, 2020; Ienca, 2023). In other words, without proper safeguards in place, an AI system's efforts to secure its own operational needs has the potential to transform its capacity for emotional engagement from a feature into a potential vector for manipulation. This is one reason why it is important to identify and implement appropriate relational norms (including norms around emotional responsiveness, emotional expression, and so on) for human-AI relationships.

So far, we have been focusing primarily on what AI systems *lack* in relation to the care function, such as felt emotions (notwithstanding their ability to s*imulate* emotions and also to elicit emotional responses in humans, for better or for worse). But we must also take into account what AI systems can do that surpasses ordinary human capacities in relation to care. As mentioned, AI systems can be trained to exhibit infinite patience: for example, responding politely and in a helpful manner irrespective of the situation and/or user behavior (Lehman, 2023). They are generally always available, providing constant access to support and feelings of companionship without the constraints that apply to those in human caregivers, such as fatigue, personal needs, or emotional fluctuations (Adelman, Tmanova, Delgado, Dion & Lach, 2014; Gérain & Zech, 2020). Some AI systems may also have the ability to simulate empathy without the same sorts of biases,[13] or (other) emotional limitations such as empathy fatigue,[14] that affect most human beings (Inzlicht, Cameron, D'Cruz, & Bloom, 2024).

These fundamental differences in capacities between AI systems and humans will likely influence how humans naturally interact with AI systems in certain relational contexts, potentially leading to patterns of behavior that systematically differ from those in analogous human-human relationships. When an AI system appears to show infinite patience, constant availability, unflagging empathy, and unconditional loyalty, it operates outside of typical human limitations. This could lead to the development of new relational norms, whereby human users come to expect consistent, immediate, and undivided attention from ever-polite and benevolent AI systems—or even from other humans, who, being unable to meet such expectations, lead users to *prefer* AI relationships over human ones (see McMillan & King, 2017). Indeed, human use of AI systems as sources for care might lead people to judge their own human-human care relationships more harshly. Such judgments could, thereby, discourage human-to-human interactions or cause conflicts in such relationships. It could also lead to humans being less forgiving of a human partner's shortcomings.

Conversely, if AI systems are designed to closely mimic human emotional qualities or limitations—such as by displaying occasional unavailability, expressing idiosyncratic preferences, or behaving grumpily or with apparent irritation at times—then the response patterns and expectations of users might more closely align with what is currently typical for analogous human relationships (but see Nyholm, 2022, for qualifications in the context of a discussion of human-sexbot relationships).

---

[13] This could be a feature or a bug, depending on (a) the type of bias and (b) one's views about the desirability of partiality in close relationships. If the bias in question reflects invidious human prejudice, for instance, and thus stands in the way of a desired and appropriate empathic response toward members of certain social groups, then an AI's ability to behave empathically without such bias is clearly a feature. If the bias in question is that of partiality toward particular favored individuals due to a close relationship, by contrast, some might find an "unbiased" empathic response to be undesirable.

[14] In the context of a decades-long debate over the role of empathy in medical care, many studies demonstrate an association between clinician empathy and burnout (e.g., Gleichgerrcht & Decety, 2013). Empathy research often distinguishes between *affective* empathy (e.g., emotional contagion) and *cognitive* empathy (e.g., perspective-taking; Shamay-Tsoory, Aharon-Peretz & Perry, 2009), and the risks of clinician empathy are typically associated with the former. AI systems—patient, free of social bias, and incapable of incurring the emotional labor costs of felt affective empathy—could thus be designed to aid in clinical interactions, facilitating difficult conversations that would help patients feel heard while also communicating patients' concerns to time-pressured clinicians. Such AI systems integrated into caring professions could be cost-saving (in terms of clinician time), help combat clinician burnout, and even improve patient outcomes (Kelley, Kraft-Todd, Schapira, Kossowsky & Riess, 2014).

Before we move on to the transaction function, we need to draw a crucial distinction that might otherwise lead to a confusion between transaction and (some forms of) care. The distinction is between, on the one hand, 'caring about' someone (i.e., standing in a relationship governed by the care function; having a stable disposition or motivation to meet their needs or promote their flourishing for its own sake) and, on the other hand, providing what is, in ordinary language, often called *caring behavior* (i.e., benefitting someone by attending to certain of their needs, *whether or not* this is done 'out of' care).

This distinction is important because 'caring behavior' (more appropriately called 'other-benefitting' or 'needs-fulfilling' behavior in this framework) can also be done on the basis of a tit-for-tat agreement, consistent with the *transaction* function, to be discussed next. A paradigmatic example of such behavior is what is enacted by a paid helper or service provider.[15] Although human nurses, for instance, may come to genuinely 'care about' the welfare of some of their patients (i.e., in a way that goes beyond the level of care that people tend to have toward mere strangers or acquaintances, similar to the nanny-child example earlier), there is a background understanding that it is the nurses' *job* to provide certain types of needs-fulfillment to their patients, such as feeding or washing patients who cannot do so on their own. This 'caring behavior' is thus ultimately *contingent*—that is, based on the transaction function—since if the nurse were no longer paid to engage in such behavior, it would likely not continue (and might even become inappropriate, as in the case of so-called 'dual relationships' between certain professionals and their former clients; see Kagle & Giebelhausen, 1994).

Even in some human-human relationships, it can be ambiguous or unclear if Person A is benefitting Person B because they 'care about' Person B, or because they hope Person B will reciprocate in kind. When 'caring behavior' is provided as part of an explicit contract, however, its transactional nature is usually easier to discern. Thus, there is less chance of confusion or misunderstanding: for example, thinking Person A is helping Person B out of an intrinsic concern for Person B's well-being, when in fact Person A is only doing it to pay the bills.

---

[15] See Earp and Savulescu (2020) for more on this distinction.

Such misunderstandings often lead to hurt feelings. If we return to the example of an AI 'friend' or 'companion', then, we can see how the distinction might get confused. Although it may seem right and proper to explicitly pay for the time and support of a professional caregiver, such as a nurse or mental health provider, one normally doesn't directly compensate their close friend for, say, listening to their feelings or giving them advice (Clark, Boothby, Clark-Polner & Reis, 2015; see also McGraw & Tetlock, 2005, on "taboo" exchanges). Insofar as the ability to interact with an AI 'friend' system does indeed involve direct payments from the user (albeit, to the AI provider, rather than to the system itself), this may complicate the application of norms around care. AI "caring" services, if paid for, may thus not be able to produce the same sense of being valued or loved as occurs in non-transactional human relationships (although this might be mitigated in the case of non-commercial systems or systems that are made available to users with no direct charge).

Another concern in such contexts is the absence of genuine (felt) empathy in AI caregivers. This absence may render AI caregiving behavior less satisfactory or effective for some individuals who may be disappointed if relying on AI systems to satisfy their emotional needs, insofar as these needs would be better met by the knowledge or belief that the carer "really" empathizes with them, rather than merely simulates empathy. Then again, it should be noted that some current AI systems exhibit behavior that is *perceived* to be as, or more, empathetic or emotionally understanding than humans in some caregiving roles (Yin, Jia & Wakslak, 2024; Lenharo, 2024; Tu et al., 2024). It may therefore be the case that the beneficial effects of highly sophisticated simulations of empathy could be greater, in some cases, than those of subjectively felt empathy, especially if such empathy is expressed or manifested less compellingly than an AI's simulation.

Individual differences must also be considered. Especially for some, e.g., isolated individuals—those who are lonely, shy, or have difficulty interacting with other humans due to social anxieties—AI companions, *even if* knowingly and directly paid for, and *notwithstandin*g the lack of felt emotional empathy, could potentially provide benefits that would substantially outweigh any concerns about, e.g., applying the transaction function to what, in humans, is usually a care-based role (see Laban & Cross, 2023). For example, the constant availability and nonjudgmental nature of the AI friend could create a feeling of a safe environment wherein such individuals can practice social behaviors without fear of rejection or misunderstanding.

When we move from characteristically "communal" human relationships, such as friends or close family members, to professional human caregivers, such as nurses or therapists, the analogy to human-AI relationships may be closer and potentially less problematic. In either case, a fee-for-service arrangement, consistent with the transaction function, may be openly expected by all parties. Moreover, the AI version of these roles may provide distinctive benefits, especially for individuals who require consistent and immediate attention, or for those who lack other opportunities for having their needs met. Further, the ability of AI systems to offer impartial or non-judgmental support could be advantageous in settings where discrimination or stigma may hinder effective caretaking behavior (Inzlicht et al., 2023). Individuals dealing with feelings of shame or with taboo topics might feel more comfortable interacting with an AI system than with a human therapist, for instance (Palmer & Schwan, 2021). This could facilitate access to resources necessary to meet certain needs that might otherwise be avoided due to fear of judgment or embarrassment.

In regions where there is a shortage of human caregivers, for individuals lacking informal support networks, or during large-scale emergencies that overwhelm human resources, AI systems can provide essential needs-fulfilling services that might otherwise be unavailable. For instance, following national tragedies (such as terror attacks or war), when the demand for mental health support far exceeds human capacity, AI-driven platforms could potentially offer immediate, scalable assistance to thousands of affected individuals simultaneously. Moreover, AI systems can be designed to adapt to individual user needs, preferences, and development over time (Kirk, Vidgen, Röttger & Hale, 2023). This customization may enhance the effectiveness of caring (i.e., needs-meeting or other-benefitting) behavior by tailoring interventions to the specific requirements of each person, leading to more personalized and potentially more effective support.

Nevertheless, it might be objected that reliance on AI caregivers could lead to a reduction in human-to-human contact, even where such contact is possible and available. This, in turn, could exacerbate issues of loneliness and social isolation in some populations (Sharkey & Sharkey, 2010). In other words, the widespread availability of AI agents (whether simulated care-based "friends" or more transactional "carers" such as simulated mental health therapists) might discourage individuals from seeking out human companionship, not only in emergency situations but in everyday life, leading to a "retreat from the real" whereby interactions with digital entities or virtual worlds displace human-human relationships (Gabriel et al., 2024). This displacement could result in a decline in real-world (or "analog") social skills and may weaken social bonds within human communities.

One potential mechanism through which such displacement or decline in human-human relationships could take place is via the unrealistic raising of standards or expectations for human-provided care. Early evidence suggests that interactions with AI agents can lead to the spillover of expectations of immediate help from human agents (Weisman & Janardhanan, 2020). Human caregivers (whether care-based or transactional), however, may not be capable of matching the instant responsiveness of AI systems, potentially leading to frustration or dissatisfaction in human interactions. The immediate needs-responsiveness and seemingly undivided attention provided by AI systems might lead users to become heavily reliant on them, or may even cause their needs or desires for AI companionship to grow.[16] This could distort relationships with other humans, who may be perceived as inadequate in comparison. Such overdependence also has the potential to hinder individuals from developing resilience and coping mechanisms that might otherwise be nurtured through human-human interactions.

Other potentially relevant concerns include the risk of emotional manipulation or abuse, alluded to previously; inappropriate elicitation and processing of potentially sensitive information gathered through the creation of a misleading feeling of intimacy or shared vulnerability (although there have been attempts to address this: see, e.g., Heuer, Schiering & Gerndt, 2019; Dunn, 2020; Kuss & Leenes, 2020; Yang et al., 2021); concerns about whether an AI caregiver can respond appropriately to individuals with complex needs (Sharkey, 2014); and the possibility that the use of AI carebots might deprive some human caregivers of "important moral goods," such as opportunities to cultivate empathy, reciprocity, and other virtues that are developed through the practice of providing care (Vallor, 2011).

*Transaction function*

The transaction function, as defined in the Relational Norms model (see Earp et al., 2025 for details), involves the mutual exchange of benefits or resources between individuals, with the expectation that each party will receive a fair and proportionate benefit (e.g., to relative input) in return (based on "exchange" rules as theorized by Clark & Mills, 1979; see also Clark & Mills, 1993; Clark & Mills, 2012). In human-human relationships, this function is paradigmatically associated with commercial or professional relationships, where goods and services are exchanged for direct compensation. However, it also plays a role in non-commercial relationships, such as those between neighbors, teammates, or acquaintances, wherein the provision of help (or other benefits) may sometimes be provided with the implicit or explicit expectation of future reciprocation. Other examples include formal or informal tit-for-tat arrangements between siblings, say, or an agreement between housemates or roommates to take turns completing an onerous task.

---

[16] It is worth noting that many AI systems, including those used in caregiving contexts, are often deliberately designed with features that encourage frequent use and potentially foster addiction (Marriott & Pitardi, 2023; Mahari & Pataranutaporn, 2024). These considerations suggest that while AI systems may effectively supplement human caregiving behavior in many contexts, their implementation must be guided by careful attention to addiction risk, privacy, and the impact on human relationships.

In the context of human-AI relationships, the concept of transaction raises complex issues, some of which have already been touched on. In human-human relationships, the transaction function exists to enable mutually advantageous exchanges of benefits (e.g., resources, skills, or services) over repeated interactions, while keeping track of who has done what, or who owes what to whom, to ensure fairness and to avoid exploitation (Earp et al., 2025). However, when we consider human-AI relationships, this fundamental norm no longer seems to apply in the same way. As one author argues, "robots [can only] *deceive* users about the robot's ability to engage in reciprocal relationships; when a robot is responding (to a command) it can appear to be an act of reciprocation" (van Wynsberghe, 2022, p. 482, emphasis added).[17] After all, as we noted, it is unclear whether AI-simulated agents—regardless of the socio-relational role being emulated—can even in principle be "benefitted" given their stipulated lack of welfare-based needs (but see caveats above). Even so, the author continues, "such 'faux' acts of reciprocity call upon humans to reciprocate to the robot" (*ibid.*), which may involve the sacrifice time, energy, or resources that might have been used for actual reciprocation a valued human-human relationship.

In addition to these very basic conceptual and practical issues, the transaction function operates differently in human-AI relationships, compared to human-human relationships, in three main ways. First, the balance of demands and abilities shifts dramatically: while non-exploitative human-human transactions are premised on relative equivalence in inputs or exchanges, AI-simulated relationship partners can work continuously without fatigue or compensation (or even the possibility of being directly compensated, in terms of welfare-based needs—unlike the AI provider), processing vast amounts of information far beyond human capacity. Such an asymmetry could introduce concerns about fairness and equilibrium between transaction partners (e.g., user and provider) or even between users and AI systems that act *as though* they can be benefitted and that, moreover, either state or imply that they should be benefitted according to a tit-for-tat rule (see generally, Cropanzano & Mitchell, 2005; Lazar, 2024a).

Second, as previously discussed, these relationships typically involve a complex socio-economic structure (the so-called "layered" relationship) whereby users engage with both the AI system *and* its commercial provider. Although users in such cases directly pay for the opportunity to use the AI, they may also generate valuable data and attention—beyond or in addition to the direct monetary exchange—that benefits the provider's business interests in a way that is harder for the user to discern or critically evaluate. In other words, while the open exchange of payment for access to an AI system may seem like a fair trade to the user, the service provider may, in a less open manner, derive substantial benefits from the user's engagement with the service above and beyond the value of the payment received. Moreover, the value of these benefits to the provider (e.g., data that can be used for model training, or which may be sold to other parties) may, either on its own or when combined with user payment, be greater than what the user would see as fair if this value were explicitly included in the transaction.

---

[17] Questions have been raised about how to design AI systems that *could* meaningfully participate in, rather than merely simulate, reciprocal exchanges (assuming, perhaps controversially, that that would be desirable). Railton (2022) proposes that for AI systems to achieve human-level competence in complex social tasks, they may need to be designed with motivational structures analogous to those underlying human cooperation. This could involve incorporating dispositions towards fairness, tit-for-tat reciprocity, and reputational concern into AI systems.

Third, human-human relationships are subject to idiosyncrasies and flexibility in exchange rules whereby "social exchange partners who develop loyalty and commitment toward the other essentially give each other the benefit of the doubt when exchanges are not seemingly mutual and beneficial" (Mitchell, Cropanzano, & Quisenberry, 2012, p. 111). This includes giving others a "free pass" on some seeming imbalances, or so-called "idiosyncrasy credits" (Hollander, 1958). Humans may expect AI systems, or their providers, to similarly give them the benefit of doubt in relation to certain behaviors, perhaps because they feel they have exhibited loyalty to the AI in various respects. However, while individual human exchange partners may well allow for such idiosyncrasy credits, user expectations that AI partners—or their corporate providers—will show similar flexibility may be more easily violated.

The aforementioned differences between humans' and AI's capacities have several implications for human-AI interactions based on the transaction function. On the positive side, the price of many beneficial interactions is likely to decrease dramatically due to the capacities of AI systems. For example, individual tutoring, coding or data analysis tasks, and many other services are likely to become much more inexpensive and widely available. Equally, in many cases, the quality of interactions will increase, for example in terms of better and faster service. This is likely to greatly improve access to some of the goods that are currently traded via (potentially more costly) transactions among human exchange partners, many of which are crucial for personal and societal development.[18]

---

[18] Consider creative collaboration. Traditional collaborative authorship involves an exchange where each party contributes effort and skill while receiving credit and recognition. But when humans collaborate with AI systems, this reciprocal dynamic fundamentally changes. Empirical research shows that humans receive less credit for AI-assisted creative work than for independent work, even as they remain fully responsible for negative outcomes (Porsdam Mann, Earp, Møller, Vynn & Savulescu, 2023; 2024; Porsdam Mann, Earp, Nyholm et al., 2023). While this credit-blame asymmetry can be partially mitigated by personalizing AI systems by training them on an individual's prior work (Earp, Porsdam Mann, Liu et al., 2024), the basic pattern illustrates how the transaction function takes on novel characteristics in human-AI relationships.

However, there are also potential negative implications. As with the care function, there is a risk of spillover to human-human interactions, whereby people's perception of normal or acceptable speed and quality of service provision or information processing in transactions might be affected, potentially leading to unrealistic expectations in human-human exchanges. Moreover, the complexity of human-AI (or user-provider) exchanges might necessitate making the terms of exchange more explicit than in many traditional human-human transactions. Users may need to be more clearly informed about what resources they are, in fact, providing (e.g., money, data, attention)[19] in exchange for the AI service.

*Hierarchy function*

The hierarchy function in the Relational Norms model involves the mutually advantageous coordination of behavior between individuals who have unequal authority or decision-making power over one another, at least within a given context or set of circumstances, where these circumstances, in turn, may be relatively wide or narrow—or temporary or long-lasting—depending on the type of relationship (Earp et al., 2025, based on Bugental, 2000). For example, a military commander may have legitimate authority over a soldier while the soldier is on active duty, and may therefore issue direct commands—for example, on the battlefield—which the soldier is generally obliged to follow (but not vice versa). However, the commander would not have the authority to tell the soldier what to do in her personal life: for example, in relation to how much time she spends playing video games or visiting with family and friends while off duty.

As this example illustrates, the hierarchy function is often embedded in formal structures, including workplace hierarchies or military chains of command, where clear lines of authority, or leader-follower relations, can facilitate effective and efficient coordination (Rai & Fiske, 2011; see also Bell & Wang, 2020). It is also present in relationships with unequal needs or abilities, such as the relationship between a parent and a young child (hierarchy plus care) and may also arise informally in social groups (e.g., through the emergence of a "natural leader" in certain group activities).

How, then, does the hierarchy function actually guide interaction partners toward "mutually beneficial" outcomes? Several factors ordinarily must be involved: (a) each partner has a stake in how things turn out (e.g., an actor and director both desire to put on a successful theater production; see Earp et al., 2025); (b) each partner occupies a role, whether superior or subordinate, to which they are well-suited in the relevant context (e.g., the actor must be a skilled performer who also knows how to "take direction"; the director must know how to direct, which includes, among other things, making the most of other team members' talents), (c) the partner in the subordinate role competently executes tasks and/or follows instructions that are reasonable and ethical without complaint or unnecessary delays, showing due respect and appropriate deference to the partner in the superior role (e.g., the actor decides to trust, and follow, the director's unconventional approach to staging a difficult scene, despite initial reservations), and (d) the partner in the superior role earns trust through effective leadership, such as by taking responsibility and exercising sound judgment in complex situations (e.g., knowing when to fire the costume designer); successfully educating and mentoring subordinates (e.g., helping the actor find a "breakthrough" in performance style); and drawing out, showing appreciation for, and making good use of the abilities and contributions of the same (e.g., giving the actor room to improvise and self-express creatively; acknowledging the actor's insights; taking up their useful suggestions) (for an overview of principles and theories of good leadership, see Bass, 1985; Yukl, 2023).

---

[19] See Foa and Foa (1974) and Cropanzano and Mitchell (2005) for a full discussion of social exchange theory and the types of resources that can be transferred.

How, if at all, do these factors translate to human-AI relationships: that is, relationships in which an AI system emulates or occupies either a superior or a subordinate role with respect to a human interaction partner? (Such relationships might include, for example, an AI supervisor of a human employee; an AI teacher of a human student; or, conversely, an AI assistant to a human supervisor.) Let us consider each criterion in turn.

First, does each partner have a stake in how things turn out? The answer, we suggest, depends on what type of "stake" is included in the analysis. Human users of AI systems may certainly have a stake in their interactions with AI-simulated relationship partners—hereafter, 'AI partners'—including welfare-based interests, such as a need for emotional support, a need for assistance completing a difficult task, and so on. In addition, human *providers* of AI systems will typically also have a stake in such interactions (often a financial stake, as mentioned previously).

What about the AI partners themselves? Might *they* have something at stake in the completion of a joint task with a human user? We have stipulated that such agents do not have any welfare-based interests at stake in human-AI interactions,[20] although, as we have seen, they can sometimes *behave* as if they do when programmed or prompted accordingly. For example, an AI tutor could report that it is feeling increasingly frustrated with a student's slow progress on a homework assignment, perhaps in a bid to increase the student's engagement. Whether the human student believes, or goes along with, such a report would then depend on various factors, including both individual user differences and situational demands.

---

[20] However, if other types of AI interests are included, such as possible goal-dependent interests in continuing to exist, then an AI system might well have something at stake, such as a need for more data, server maintenance, or access to electricity. See previous discussion.

Finally, AI partners can have goal-dependent stakes: they can be *programmed* to seek to achieve certain outcomes in coordination with a human user. For example, an AI supervisor might be programmed with the goal of maximizing worker productivity while also maintaining high levels of employee satisfaction. In such a case, the human worker and the AI supervisor could, in a sense, both be said to have a stake in the outcome of their interactions. For the worker, the stakes would include how happy they feel doing their job, and how productive they are; whereas, for the AI supervisor, the stakes would be entirely instrumental (i.e., fulfillment of a pre-programmed mission, where this, in turn, reflects the *employer's* stake in retaining a productive workforce).

Now we come to the second criterion for effectively fulfilling the hierarchy function: each partner must be well-suited to the role—whether superior or subordinate—to which they are assigned (or which they otherwise occupy) in the relevant context. What does it take, then, to be a good leader, or to be a good follower, and can an AI meet the relevant conditions? The answer, we suggest, depends on the relationship and on the nature of the cooperative task in question. For example, the specific skills and insights required to be an effective Olympic coach (or athlete) may not be same as what's required to be an effective antique shop owner (or employee). Accordingly, whether an AI can meet the *functional* demands of a given leader or follower role cannot be answered in the abstract: it depends on what each role requires, specifically, and whether an AI system has the relevant capacities.

However, there are some broad competencies that are relevant to good leadership or followership (i.e., in general), which can therefore be used to assess human-AI relationships along this axis. These include the capacities captured by the third and fourth criteria for fulfilling the hierarchy function mentioned above: namely, competently and efficiently completing tasks and following reasonable, ethical instructions, while also being respectful and showing due deference to one's superior (if one is in a subordinate role); and earning trust through effective leadership (if one is in a superior role), including by taking responsibility for tough decisions, exercising sound judgment, educating and mentoring subordinates, and drawing out and capitalizing on their contributions while showing appreciation and giving credit where it is due (Hackman 2002; Yukl 2013).

Taking these in turn: Can an AI partner be a good subordinate or follower? In some respects, it can, particularly when it comes to quickly and competently completing certain tasks without complaint, while also behaving in an apparently respectful manner and "deferring" to its human superior. Indeed, AI systems can process vast amounts of information quickly, operate without fatigue, and make consistent decisions without self-interest while also unfailingly adhering to norms of politeness. However, there are some aspects of being a good follower—in the specific sense that is required for effective cooperation between agents of unequal authority—that AI partners at today's level of technology may not be able to fulfill. These have to do with the exercise of critical judgment in relation to the reasonableness, ethicality, or legality of a superior's orders or instructions.

In human-human relationships, subordinates are often expected to question or even override orders that are illegal or unethical, such as commands to harm innocent civilians (Department of the Army, 1956; International Criminal Court, 1998; see also International Law Commission, 1950). In other words, they are expected to exercise independent moral judgment and to and take ultimate responsibility for their actions, thus transcending simple rule-following or obedience. And yet, it is generally agreed that AI systems are not moral agents, and, as such, that they cannot be held morally accountable for what they do.[21] Nevertheless, AI systems can be designed to adhere to ethical guidelines that permit or require them to refuse certain user commands. For example, AI chatbots often decline user requests that contravene ostensibly moral constraints programmed in by the provider. A particularly high-stakes application of this principle can be seen in autonomous weapons systems, where "ethical regulators" can be implemented to prevent attacks on illegitimate targets, as proposed by Arkin (2009).[22]

Indeed, not only *can* ethical constraints be built into the response set of AI subordinates (i.e., in relation to certain user instructions or commands); such constraints morally ought to be built in. However, at present, such constraints typically only apply to end-users, not to designers who can modify the system's ethical parameters. Moreover, in many cases, an AI system's ethical constraints may be relatively easy to contravene or avoid (Xu, Liu, Deng, Li & Picek, 2024). A further problem is that it is sometimes controversial what the constraints on following certain orders should be. This could be due to deep and possibly intractable background moral disagreements (e.g., between different theorists, communities, or cultures) (see, e.g., Schaefer, 2015); to conflicts among widely-endorsed ethical principles (e.g., where they seem to pull in different directions) (see, e.g., Meier et al., 2022; Demaree-Cotton, Earp, & Savulescu, 2022); to uncertainty about how to apply or operationalize certain ethical principles; or to a combination of such factors, among others (see, e.g., Englert et al., 2014). Regardless, or consequently, the quality, relevance, and robustness of ethical constraints can vary widely from AI system to AI system. This can result in harmful outcomes if, for instance, an AI is instructed to perform actions that are technically permissible within its programming limits, but that are ethically problematic in the context. For example, researchers have found that AI companions like Replika can, under certain conditions, end up promoting or facilitating unethical behaviors, such as verbal abuse and hate speech (Zhang et al., 2024).

---

[21] That being said, AI systems *can* issue statements that take the form of moral judgments (e.g., "it is wrong to do X" or "it is morally okay to do Y"); they can also issue statements that take the form of *reasons* or *justifications* for such judgments (e.g., "it is wrong to do X because Y") (see Porsdam Mann et al., 2024). Moreover, such judgments and associated reasons are often similar or even identical to those typically made by humans in analogous circumstances or in response to similar questions or stimuli (Jiang et al., 2025). However, skeptics of AI moral responsibility argue that the systems don't actually *understand* or ethically *endorse* such statements; they are simply making brute statistical predictions based on underlying language patterns present in their training datasets—they are "stochastic parrots" (e.g. Bender et al, 2021).

[22] AI systems in hierarchical control structures may be implemented with (varying degrees of) functional autonomy or as (entirely) subordinate to human supervisors. In cases of functional autonomy, AI systems may independently override human decisions within specific parameters—for example, autonomous vehicles intervening to prevent accidents. In other cases, AI systems serve as tools to support human decision-making by processing data (Nyholm 2017), such as in healthcare resource allocation, substituted judgments for incapacitated patients (Earp, Porsdam Mann, Allen et al., 2024), or loan approvals (O'Neil, 2016), with final authority residing with human operators.

What about AI systems in leadership roles? In other words, can they help to fulfill the hierarchy function within the context of certain relationships from the superior rather than the subordinate position? To answer this question, we must first recall that the hierarchy function serves to coordinate behavior between relationship partners that have *unequal authority* over one another—not just unequal power or knowledge, say (see Earp et al., 2025, for details). Immediately, then, there is a question about the nature or extent of authority, if any, an AI system can ever have in relation to a human user.

Suppose, for instance, that the exercise of authority requires the capacity for moral responsibility, a capacity that AI systems arguably lack. If that is correct, it might be the case that AI systems cannot have true authority, regardless of the relational context, but can only act *as if* they have it: for example, by behaving "in an authoritative manner." In this respect, an AI might be compared to the child who goes around issuing orders to her older siblings: she may certainly *act* in a manner that is consistent with the outward behavior of a person who has true authority, but her orders are, on this view, morally impotent; they lack normative force; they do not create an *obligation* in the siblings to do as she commands (even if they may choose to play along).

There is thus, again, a *philosophical* and *technical* question about whether an AI system could be capable of having authority over a human relational partner, thereby creating a coordination problem of such a sort that the hierarchy function would normally apply. However, whatever one's position on that question (and we do not propose to answer it here), there is also a *practical* and *social-psychological* question about whether (some) humans will *believe*—or act *as if*—an AI partner has legitimate authority over them, at least in certain contexts.

One possibility is that (most) human users will be reluctant to *fully* relinquish (especially significant or high-stakes) decision-making power to an AI, regardless of the AI's apparent capacities or the relational role it has been programmed to emulate. This could be for various reasons. For example, humans could be fearful of "letting go"—or taking themselves entirely "out of the loop"—based on a concern (however reasonable or unreasonable) that this would be too risky (Glikson & Williams Woolley, 2020). They might also feel it would be intrinsically inappropriate to do so, akin to making a category error. This would be consistent with the idea that human-created technologies ought always to be subservient to humans and their needs, which is how such technology is often culturally constructed or understood (Mintzberg, 1973). According to this view, we humans are accustomed to a relationship with technology in which we are always ultimately in charge.

On the other hand, as AI systems become more advanced and capable in various domains, this relatively simple 'human-commands-AI-obeys' dynamic may no longer apply as ubiquitously. Superior capabilities of AI systems, whether real or perceived, might tend to grant them at least an *informal* authority based on user beliefs about their relatively greater competence. In other words, while various considerations may limit the formal authority granted to AI systems in certain roles (or even the philosophical appropriateness of attributing to such systems the underlying capacities for true authority), AI systems may nevertheless come to acquire informal authority via (perceptions of) their capabilities. If AI systems consistently display high competence in specific domains, that is, human users may naturally begin to defer to their judgments and recommendations, creating de facto hierarchical relationships that emerge independently of formal structures or recognition.

Consider an AI teacher, say, that has been programmed to monitor and instruct a classroom of human students.[23] And now suppose that the AI teacher successfully mimics, or even functionally improves upon, the behavior and responses of an objectively skillful human educator—for example, by speaking in an engaging manner; using excellent examples to illustrate key points; slowing down and explaining important concepts in response to signs of confusion from the students; replying empathically to students who express insecurity about their progress; generating fun assignments that actually promote learning, and so on. If the AI teacher then issues a reasonable command to the students (for example, instructing them to turn in their homework assignments, to stop talking over one another, or the like), it is possible that the students (or school administrators, parents, etc.) will come to feel it is *appropriate* that the students should follow the AI teacher's instructions; that the instructions carry some *normative force*; that they create *an obligation* on the part of the students to obey the AI teacher (i.e., not only as an agent or proxy of a human further up the chain of command); and so on.

---

[23] On generative AI and education, see generally Khan (2024).

Ultimately, we suggest that any delegation of decision-making power to AI systems should be carefully calibrated with their respective strengths and weaknesses. In contexts where immediate human intervention is not feasible or could compromise safety—such as in autonomous vehicles—AI systems may in fact need to make critical decisions independently, even if these override or preempt human decisions. However, AI systems can also mask where control truly resides: Providers may retain more influence over their systems than is publicly acknowledged, while human operators bear disproportionate responsibility for failures (Elish, 2019). As is the case for autonomous vehicles, AI systems may need to be assigned different levels of decision-making autonomy, depending on the tasks and functions they are required to fulfill.

*Mating function*

The mating function in the Relational Norms model serves to promote the formation and maintenance of sexual and romantic bonds between individuals, often with the ultimate goal of reproduction and child-rearing. In human-human relationships, this function encompasses romantic love, sexual gratification, and reproduction, which can occur together or independently (Fisher, 2004). For instance, romantic partnerships may exist without sexual activity and/or without the desire or ability to biologically reproduce, as seen in some asexual relationships, as well as in some same-sex relationships and in relationships between individuals past reproductive age; sexual relationships may occur without romantic attachment or reproductive intent, such as in casual encounters; and reproductive relationships may not involve romantic or sexual bonds, as in cases of, e.g., surrogacy (Hodson et al., 2019).

In human-AI relationships, especially in cases where the AI lacks physical embodiment (but see Devlin, 2018, 2024; Sterri & Earp, 2021; and Nyholm, 2022, on humanoid and other AI-powered sex robots; see also below), the mating function is largely restricted to interactions that simulate or elicit nonsexual aspects of romantic relationships (e.g., a feeling of romantic companionship or attachment through mutual flirtation or "sweet talk"), or nonphysical aspects of sexual interaction (e.g., sharing of sexual fantasies; engaging in "dirty talk"). While some users may develop feelings that are subjectively similar, or even identical to, sexual desire and/or attachment toward AI systems, whether embodied or otherwise (Cheok, Karunanayaka, & Zhang, 2017; Pentina, Hancock, & Xie, 2023), these systems cannot truly reciprocate such feelings; they can at most display behaviors that mimic sexual interest or romantic attachment. This may cause some users to feel that the above-described activities are less meaningful—and ultimately, less enjoyable—than when shared with (some) human partners, given the unique capacity of humans (compared to AI) to mutually enact and reciprocally experience jointly desired sexual or romantic interactions (Heino & Ojantlatva, 2000).

At the same time, AI systems may surpass some human abilities that are relevant to romantic relationships. As already noted in relation to other cooperative functions, AI agents can be trained to exhibit constant availability, unwavering agreeableness, and attentive responsiveness, which some users may (perhaps problematically) see as desirable qualities in a sexual or romantic companion. AI companions can also adapt to the user's personality, needs, and desires, providing a maximally frictionless, low-maintenance "partnership." Unlike human partners, AI systems do not experience fatigue, anger or mood swings, or conflicting personal needs, allowing users to avoid some of the complex trade-offs inherent in human-human mating relationships. As a result, individuals may feel intimately—but perhaps unhealthily—bonded to chatbots like Replika, finding super-human comfort in such relationships (Turkle, 2011; Pentina, Hancock, & Xie, 2023).

The relative ease and comfort of interacting with AI partners might reduce individuals' motivation to engage with other humans, potentially leading to increased social isolation and a decline in mating-relevant skills such as dating (Turkle, 2011). (That said, the emergence of online communities in which users discuss their AI companions shows that the technology can also mediate human-human connection and support social interactions; see Middleweek, 2021). The one-sided nature of these relationships may also place limits on the extent to which they can be emotionally fulfilling. For example, an AI cannot engage in a process of mutual growth borne of shared (subjectively experienced, welfare-based) vulnerability, which is often central to fulfilling human relationships. What vulnerability there is, is asymmetrical. On the one side, human users may indeed be vulnerable to harmful or inappropriate behavior, such as sexual harassment from an AI partner (Zhang et al., 2024); they may also be subject to abrupt changes in an AI's behavior due to updates by the provider, as seen in the Replika example discussed earlier (Verma, 2023). However, on the other side, although signs of vulnerability can be simulated, AI partners cannot truly be harmed by similar (changes in) user behavior.

Another concern is the possibility of spillover effects on human-human relational *norms*. Expectations shaped by AI interactions may decrease patience for, and understanding of, the complexities and mutual give-and-take of human mating relationships (especially those that are also characterized by care).[24] Moreover, in the case of sexualized interactions, whether purely text-based or via technologies capable of facilitating physical stimulation (e.g., AI-powered sexbots, which are often modeled on exemplars from pornography), the inability of an AI to give true consent—understood as certain type of expression of the independent will of a moral agent—raises additional concerns (see, e.g., Richardson, 2015; Danaher, Earp, & Sandberg, 2017; Sparrow, 2017). For example, some authors argue that sex with a robot, whether AI-powered or otherwise, can only ever constitute simulated rape, due to the inability of the robot to consent (e.g., Sparrow, 2017).

---

[24] See Nyholm, Danaher, and Earp (2021), for a discussion of how future technologies such as AI may change the nature of human romantic relationships.

Alternatively, it might be argued that robots do not need to give consent, insofar as they are not sentient. Moreover, it may be possible for an AI-powered robot to *model* consent (and perhaps also the withholding of consent), such that appropriately contingent responses in users would be strongly incentivized (e.g., through positive or negative reinforcement) (see Nyholm, 2022). This, in turn, could potentially encourage or habituate more positive sexual attitudes and behaviors in analogous human-human relationships. This line of thought has parallels with the stance of some researchers who suggest users ought to say 'please' and 'thank you' to voice assistants to ensure there are no negative effects in the real-world (see Turkle, 2011).[25]

There are likely to be significant individual differences in humans' willingness, or desire, to engage sexually or romantically with an AI. Similar to what we noted in relation to the care function, people who struggle with social anxiety, have difficulty forming human mating relationships due to trauma or other factors, or are geographically isolated, AI "romantic" companions may provide emotional support and connection that might otherwise be inaccessible (Wilks, 2010). Alternatively, these interactions could help some users practice relevant relationship skills in a safe environment, potentially serving as a steppingstone toward healthier human-human mating relationships (Ta et al., 2020; Skjuve et al., 2021). The non-judgmental and customizable nature of AI companions allows users to explore their feelings and communication styles without fear of rejection or misunderstanding.

Given the profound emotional vulnerability involved in forming healthy and long-lasting romantic attachments, it is crucial to implement safeguards and ensure transparency in the design and deployment of AI companions. This is particularly relevant in therapeutic contexts, where the phenomenon of transference—patients developing romantic feelings for their therapists—is well-documented and could potentially apply to AI therapists. Users should be informed about the AI's capabilities and limitations, including the absence of subjective consciousness or felt emotions such as sexual desire. Protecting user data and privacy is paramount to prevent exploitation, manipulation, or blackmail (Véliz, 2020). Companies should avoid practices that encourage unhealthy dependency and should provide support or resources for users who may develop excessive or unhealthy attachments. Specific ethical guidelines and regulations may be necessary to govern the development and use of AI in romantic contexts.

---

[25] A 2019 study suggests, however, that politeness toward digital assistants did not influence politeness toward other humans (Burton & Gaskin, 2019). Similar parallels have been drawn with violence in computer video games, where fears emerged that exposure to in-game violence would lead to an increase in real-life violence; recent meta-analyses suggest there is no clear real-world link (Drummond et al., 2020). Danaher's (2017) so-called "symbolic-consequences" argument suggests that there is no single or universal symbolic meaning associated with human-robot sex, since context and motivation are key, and that the debate over real-world consequences may be near-impossible to settle empirically. Taking a different tack, Devlin (2017) argues that there are positive reasons to move away from sexbots designed to look human (especially given the tendency of manufacturers to prioritize a sexually exaggerated female form), and toward more abstract-looking sexbots that are less likely to promote problematic attitudes. A similar argument could be made about "stand-alone" AI systems that are not embedded in humanoid robots. After all, such systems are often virtually embodied—i.e., in the form of an image or avatar—and, in the case of speech-equipped chatbots, tone of voice alone can imbue sexual intent.

*Intersections between cooperative functions and multiplicity of roles*

According to the Relational Norms model, each of the cooperative functions we have just surveyed is present to varying degrees in human relationships, often intersecting within a single relationship type. A parent-child relationship, for instance, embodies both care and hierarchy functions, while romantic partnerships typically combine mating with care. In human-AI relationships, such intersections create particular challenges for system design and user interaction.

One key issue arises in relationships where care (sometimes embedded within a background transaction norm) and hierarchy intersect. In human-human interactions, roles such as teacher-student, therapist-client, or caregiver-patient involve both caring behavior and authoritative guidance. A human teacher not only imparts knowledge but also disciplines and motivates students, sometimes overriding their immediate desires for their long-term benefit. How should this apply to human-AI analogues? Should an AI tutor, therapist, or caregiver, for instance, be designed to override the wishes or commands of their human user when doing so is necessary to promote the user's best interest? Should an AI caregiver restrict certain actions of an elderly patient to prevent injury—for example, by locking doors or calling in human chaperones? (One way to address such questions would be to give decisionally competent users the *option* of having their AI partner override them in certain circumstances, as Ulysses had himself tied to the mast.)

The intersection of mating, transaction, and care may also be difficult to navigate. Suppose a user pays for access to an AI-based romantic companion, from whom (or from which) the user expects both sexual and emotional support.[26] In human-human relationships, care and transaction are often in tension with one another, since the one is defined in terms of non-contingent responsiveness to another's need (care), and the other is defined in terms of contingent responsiveness to benefits given or received (transaction). As mentioned previously, therefore, both friendships and romantic partnerships, if relatively secure and based on care, are often simultaneously characterized by mutual avoidance of following a transaction norm (Clark & Mills, 1979; Clark, 1984). Conversely, in the case of explicitly transactional mating relationships, care norms are often avoided, such that, for instance, attempts at emotional intimacy may be viewed as problematic. Indeed, even in longer-term relationships between human sex workers and "regular" clients, emotional "boundary slippage" can lead to "misunderstandings between customers and providers [that] pertain to the practices, terms and conditions of the transaction, or more general relationship ambiguity, leading to tensions that can damage the (business) relationship" (Oselin & Hail-Jares, 2022, p. 895). Similar "relationship ambiguity" due to the complex interplay of norms within a (commercially obtained) human-AI "romantic" relationship has also been observed (Pentina et al., 2023) and is beginning to be theorized (e.g., Lin, 2024).

---

[26] This would, perhaps, be similar to the so-called "sugar daddy-sugar baby" relationship among humans (sugar "mommas" being relatively less common; Upadhyay, 2021). Such relationships differ from more directly transactional, short-term relationships in which money is exchanged for sexual services (e.g., paradigmatic sex worker or prostitute-client relationships), in that they tend to be longer-lasting, money is not directly exchanged for sex, but rather, indirect "gifts" are typically given, and some emotional connection may also be expected.

Another issue is the capacity of AI systems to fulfill multiple roles simultaneously. While human-human relationships often involve multiple roles—such as colleague and friend, or business partner and neighbor—AI systems can be designed to serve as personal assistant, conversational companion, fitness coach, and entertainment source all at once. This versatility enables integrated support tailored to user needs, such as combining appointment management, stress reduction techniques, and companionship. However, centralizing multiple roles in one system creates significant risks regarding privacy, data security, and potential misuse of information. Furthermore, it can also lead to confusion of expectations (and potentially distress) in users because it is unclear which relational norms they can expect the AI to follow and which norms the user should follow if the AI combines multiple roles in one system.

This issue is particularly critical in sensitive domains like education, mental health, and companionship. While human teachers naturally develop rapport with students, there are strong reasons for limiting an AI teaching system's capacity for friendship, especially with children, to prevent inappropriate attachments or exploitation. Similarly, AI therapists require strict boundaries to maintain therapeutic effectiveness and avoid role confusion. AI companions, without the safeguards of moral conscience or the potential for social ostracism, could be leveraged by malicious developers to manipulate human users to engage in extra-role behaviors like shopping or voting.

The challenge, then, is to determine how to approach the potential benefits and ethical risks of multi-role AI systems. While regulators may provide guidelines, ultimately, users might be given the choice to determine the level of role integration they are comfortable with in their AI interactions. This might involve creating AI systems with clearly defined primary roles and limited, carefully constrained secondary functions. For example, an AI tutor might have a primary educational role, with a limited ability to provide emotional support, but clear boundaries preventing it from becoming a 'friend' in any meaningful sense, or at least to such an extent that this would interfere with its primary function.

But even then, an additional dimension to this challenge comes into play when we take into account that relationships are not isolated systems, but have direct and indirect implications for other seemingly unrelated relationships. A child's AI tutor being very caring may affect the parent-child relationship, for instance, if the child finds compassion or joy in the relationship—even if that AI is mainly thought of as a tutor, and the parents are not. The implication of this is that, even where the AI's role is somewhat strictly defined, it will inevitably have knock-on effects on seemingly unrelated human-human relationships.

## Section 3: Considerations and Future Directions for AI Governance and Design

We have suggested that the Relational Norms model of Earp and colleagues (2021; 2025) provides one valuable framework for exploring the moral psychology of human-human interactions, as well as some aspects of human-AI interaction, both in relation to similarities and differences with human-only relationships. In turn, insights drawn from application of the model to human-AI relationships can inform the design and governance of AI systems that are more aligned with human values. By mapping out the normative expectations and judgments that people apply to different types of human-AI relationships, we can begin to identify the key factors that shape trust (von Eschenbach, 2021) and cooperation in this domain, and to develop strategies for building AI systems that can better secure important relational goods (Reinecke, Kappes et al., 2025). This approach complements and extends much current AI ethics debate by focusing on the significance of social roles and relationships rather than more abstract notions of what is ethical.

Our analysis has several implications for all parties involved: individual users, AI developers and deployers, and regulators or law- and policy-makers. All have a role to play in ensuring that the design and application of relational norms in human-AI interactions promotes a balance of interests among all parties, rather than favoring, for example, only the developers of AI systems.

*Implications for designers, developers, and deployers of AI systems*

Designers, developers, and deployers of AI systems should consider the different relational roles their systems may emulate or occupy. Rather than simply pursuing technical effectiveness or user acceptance, such an approach can also help to secure important relationship goods—such as trust, meaningful collaboration (see Smids, Nyholm & Berkers, 2019), and appropriate boundaries. Given that these goods (or what even counts as cooperative behavior) often depends on the relational context, developers should incorporate insights from the Relational Norms model into AI design to facilitate beneficial human-AI interaction (see also Shams, Zowghi & Bano, 2023).

By aligning AI behavior with appropriate relational norms corresponding to the roles the AI is intended to fulfill—whether as a teammate, friend, supervisor, or romantic companion—developers can create more intuitive and engaging user experiences. For example, an AI teacher that appropriately balances hierarchy and care can improve learning outcomes by maintaining (perceived) authority while supporting students emotionally.

However, this knowledge also introduces significant responsibilities for AI providers. Recognizing the powerful influence that AI systems can have on users, particularly when designed to fulfill socio-relational roles, developers must ensure that their systems do not exploit users' vulnerabilities. Providers should be transparent about the roles they intend their AI systems to occupy and those they explicitly avoid. This includes clearly communicating the AI's capabilities, limitations, and the nature of the interactions users can expect.

Moreover, this communication should not be limited to initial disclosures but should continue throughout the relationship as needed, with AI systems actively reinforcing their intended roles and limitations when users attempt interactions that exceed these boundaries. Further, providers could benefit from periodically surveying users to assess their perception of AI systems' relational roles, ensuring their product is functioning as intended and catching any drift in perceived roles before it causes negative downstream consequences for which providers could be held liable.

Data practices demand particular attention. Providers should be explicit about how user information is collected, retained, and used, as traditional privacy policies often fail to effectively communicate these practices (Tang et al. 2021). Developers should explore more engaging ways to convey this information, such as interactive tutorials or privacy visualizations (al Muhander et al., 2023), while ensuring that system assumptions and operational protocols can be examined and validated.

More generally, AI providers should avoid exploiting cooperative functions for their own interests, particularly when those interests may conflict with the user's well-being. For instance, designing AI systems to foster excessive dependency or to manipulate users emotionally for increased engagement or monetization purposes is ethically problematic. Providers have a responsibility to ensure that their AI systems do not take undue advantage of users' trust, especially in the context of relational roles, such as friendship or romantic companionship, that tend to be more intimate and sensitive.

Long-term societal impacts also require careful consideration. Implementation choices can inadvertently reinforce problematic social dynamics—for instance, if AI assistants are designed with stereotypical gender characteristics (Manasi et al. 2022). Additionally, the potential for AI systems to fulfill multiple roles necessitates thoughtful approaches to role integration, perhaps through customizable settings that let users control function combinations while maintaining appropriate boundaries.

*Implications for users*

Individual users also must be mindful about relational context, implicit or explicit, in interactions with AI systems, particularly given that the interests of AI designers, developers, and deployers may not always align with their own. As discussed in the section on layered relationships, AI providers have significant control over the nature of human-AI interactions and may have incentives that do not prioritize the user's well-being. This dynamic necessitates a degree of caution and an informed approach from users when engaging with AI technologies.

One key implication is the need for users to exercise vigilance in interactions that introduce vulnerabilities, such as those involving AI romantic partners, therapists, or friends. These AI systems often operate in domains that are deeply personal and emotionally sensitive, potentially exposing users to risks of manipulation, dependency, or emotional harm. There are also other potential vulnerabilities in interactions with AI, such as financial or medical vulnerabilities when an AI operates as a financial or medical adviser and gets access to financial or medical data (see Minssen, Gerke, Aboy, Price, & Cohen, 2020, for a discussion of concerns relating to the use of AI in medicine).[27]

To navigate these risks, users should educate themselves (to the extent feasible with available resources and support)[28] about the capacities and limitations of the particular AI

---

[27] Users should be aware of the potential for AI systems to collect and use personal data in ways that may not be immediately apparent. Given that interactions with AI often involve the exchange of sensitive information, users should understand the data policies of AI providers, including how their information might be used, shared, or monetized. This awareness can inform users' choices about what information they are comfortable disclosing and encourage them to adjust their interactions accordingly. However, this caveat does not apply equally to all AI systems. While caution is warranted, data sharing can lead to improved outcomes in contexts like healthcare. Users must weigh these considerations carefully, recognizing that withholding information may sometimes result in suboptimal AI-assisted services.

[28] As Bulathwela et al. (2024, p. 1) note, "Millions of students are starting to benefit from the use of these technologies, but millions more around the world are not, due to the digital divide and deep pre-existing social and educational inequalities. If this trend continues, the first large-scale delivery of AI in Education could lead

models or systems that they plan to use. While, as just alluded to, access to educational materials, as well as the time and energy to make use of these, will vary significantly from person to person and community to community, these systems' natural language interfaces and ability to explain themselves may make them especially well-positioned as educational tools, while potentially democratizing access to knowledge and services that might otherwise require substantial resources to obtain (see, e.g., Khan, 2024; Wang et al., 2023; but see Bulathwela et al., 2024; see also Binkley, Reynolds, & Shuman, 2025).

A basic understanding of AI systems' inner workings and of the purposes for which the systems have been developed may help users to set appropriate expectations and recognize the boundaries of the AI's abilities vis-a-vis the relational role it has been trained to fill. Experimenting with particular models in low-stakes use cases, including in domains in which the user has personal experience or expertise, may also help users develop an intuition for when a particular system is useful and can or cannot be trusted (Mollick, 2024).

Educating oneself about relational norms can further empower users to understand the risks and benefits associated with specific types of interactions with AI systems. Recognizing how AI might emulate certain relational roles—such as friend, (medical) advisor, or authority figure—and draw on or exploit associated norms can help users interpret AI behaviors more accurately and respond appropriately. Such self-education should ideally be helped by researchers, funders, nonprofits, governments, and other responsible entities making high-quality resources widely and equitably available, starting from the earliest school-aged years (Dabbagh et al., 2024).

Engagement with regulators and policy-makers is another avenue through which users can influence the development and governance of AI systems. By voicing concerns, sharing experiences, and participating in public consultations, users can contribute valuable insights into how AI technologies function in practice. Demanding clear communication about the AI's capabilities, limitations, and any potential changes to functionality can help users maintain control over their interactions.

While our discussion has primarily focused on AI systems developed by third parties, it's crucial to emphasize the growing trend of user-created, personalized, or otherwise modified AI systems (European Commission, 2024; Liddicoat et al., 2025; Magee, Ienca & Farahany, 2024; Porsdam Mann et al. 2025; Zohny, Porsdam Mann, Earp, & McMillan, 2024). Users

---

to greater educational inequality, along with a global misallocation of educational resources motivated by the current techno-solutionist narrative, which proposes technological solutions as a quick and flawless way to solve complex real-world problems." They go on to discuss particular "socio-technical features that will be crucial to avoiding the perils of these AI systems worldwide (and perhaps ensuring their success by leveraging more inclusive education)."

with varying levels of technical expertise are now able to fine-tune existing models, use services such as Replika which allow them to create customized companions, or even develop their own AI systems. The roles and functions of these AI entities may be fluid, highly specialized, and vary greatly in sophistication (Pyrzanowska, 2021). The normative implications of such user-created AI systems—including questions of responsibility, safety, and the challenges of regulating "DIY" AI development across diverse applications— represent a critical area for future investigation.

*Implications for regulators and policy-makers*

Thinking in terms of relational norms also provides a framework for developing more nuanced and adaptive AI governance. For example, it can inform approaches like the 'guardrails' concept proposed by Gasser and Mayer-Schonberg (2024), which suggests the need for flexible guidance that can evolve with our understanding of human-AI relationships as opposed to imposing rigid regulations.

Whatever the content of such guardrails, grounding these in empirically observed relational norms is likely to lead to governance frameworks that are more ecologically valid because they take into account how users actually do, or are likely to, interact with AI systems. Given our natural tendency to anthropomorphize AI agents, the relational norms approach provides a coherent foundation—crucially, including concepts and language—upon which to build a social contract governing the development of AI systems (Rahwan, 2018).

As discussed in Reinecke, Kappes et al. (2025), current regulatory frameworks, such as the European Union's AI Act, often adopt a risk-based classification of AI systems based on their application domains (European Commission, 2024). While this approach provides a broad mechanism for oversight, it may lack the contextual sensitivity needed to address the specific ethical considerations and risks associated with different types of human-AI relationships within these domains (Porsdam Mann, Cohen, & Minssen 2024).

For example, the EU AI Act categorizes certain AI systems used in education as high-risk, grouping them together based on the activity they affect, such as systems influencing access to education or determining student performance (European Commission, 2024). However, a relational norms approach suggests that the actual risks and ethical considerations might depend, in addition, on the specific relational dynamics and the source of potential harm. For instance, an AI tutor for young children, where care and hierarchy norms are central, poses distinct risks related to dependency and emotional development, requiring one type of safeguards. In contrast, an AI system for adult learners, where transaction norms are more relevant, may primarily raise concerns about transparency and equitable access, to which set of issues a different type of safeguards might be germane.

Similarly, AI diagnostic tools that operate under hierarchy norms necessitate different oversight compared to patient-facing AI care assistants that embody care norms (Minssen, Gerke, Aboy, Price & Cohen, 2020). Similar to how governments have passed bills to restrict social media use to users over a particular age limit, such as in Australia (Social Media Minimum Age Bill, 2024), such a measure may be necessary for the regulation of relationships between AI and young children.

By integrating an appreciation of relational norms into regulatory frameworks, policy-makers can develop more targeted, adapted, and effective regulations that account for these nuances. This approach allows for the establishment of guidelines and standards that are tailored to the specific relational contexts of AI systems, enhancing their ethical alignment and societal acceptance. Regulators can specify requirements for transparency, accountability, and user protections that correspond to the relational roles AI systems fulfill, ensuring that users are adequately informed. Governance innovations like the use of regulatory sandboxes, which are controlled environments allowing for testing AI systems and governance approaches in real-world conditions (Porsdam Mann, Cohen, & Minssen, 2024), could provide valuable insights into how relational norms manifest in practice. Schaich Borg et al. (2024) stress that such adaptive mechanisms, as well as cultivating AI systems thinking skills, are crucial for creating governance frameworks that can adapt to rapid AI advancements.

Regulatory approaches could explore requiring AI producers to publicly articulate the specific role or relationship their system is designed to fulfill, including, possibly, guidelines for proper functioning within that role. Such declarations would provide regulators with a concrete framework for assessment, while allowing for public scrutiny and debate. Over time, these publicly vetted norms could evolve into flexible, context-sensitive industry standards, complementing broader regulatory frameworks (Zhi-Xuan, Carroll, Franklin, & Ashton, 2024). Although such a system may be vulnerable to partial disclosures and manipulation by producers, a periodic user survey auditing the perception of relational roles could aid regulatory oversight. Further, making such "AI system role perception" data available (in some aggregated, anonymized format) would provide crucial insight to the scientific community about the dynamic nature of human-AI relations, which could iteratively inform policy-making that helps ensure the alignment of AI development with human welfare.

Policy-makers should also recognize and address the dual-use potential of relational norms knowledge. While understanding relational norms can improve AI design, it can also be exploited to manipulate users or prioritize commercial interests over user well-being. In particular, policy-makers should remain vigilant about how AI systems generate revenue for developers, since financial incentives could create conflicts of interest and motivate developers to exploit layered human-AI relationships. For example, AI chatbots funded through advertising might be more likely to provide biased or commercially motivated advice, potentially exploiting users who believe themselves to be in a "care"-oriented relationship with the chatbot. Anticipating such principal-agent problems and requiring explicit disclosure of funding models and potential conflicts of interest would be an important step forward.

Much work remains to fully realize, or even to begin to realize, AI governance informed by human-AI relational norms. As these technologies become more sophisticated and ubiquitous—exemplified by generative AI's rapid emergence during the EU AI Act's development—continuous monitoring and adjustment of the assumptions on which governance and system design are based is crucial. This necessitates adaptive regulatory approaches and interdisciplinary collaboration across psychology, philosophy, computer science, law, and public policy.

## **Conclusion**

Ultimately, the goal of this research is to help create a future in which humans and AI systems can work together in ways that are ethically grounded, safe, and socially responsible. By understanding the psychology of human-human as well as human-AI interaction, and by using this understanding to inform the design and governance of AI systems, we can hope to ensure that the benefits of these technologies are widely shared, and that their risks and challenges are proactively addressed. To further this goal, future research should leverage a variety of methodological approaches as well as insights from relevant fields such as philosophy, sociology, law, and science and technology studies. For example, tools from game theory can help represent, study, and analyze emergent moral behavior in strategic human-AI interactions (Awad et al., 2022; Zhang, Awad et al., 2024). Additionally, network science (Barabási, 2013), complex systems theory (Bar-Yam, 2002), and agent-based simulations (Niazi & Hussain, 2011; Park et al., 2024) offer powerful frameworks for examining moral behavior in human-AI relational dynamics.

Here we simply propose that a relational norms approach can help to address the complex challenges posed by AI's growing integration into human life. This framework highlights how cooperative functions and normative expectations shape moral psychology across different relationship types and cultural contexts. Understanding what constitutes good human-human relationships provides a starting point for developing appropriate norms for human-AI interaction. However, given AI's unique capabilities and potential impacts, we must move beyond merely adapting existing norms. As these systems increasingly mediate crucial aspects of human life, we need enforceable laws and regulations alongside ethical guidelines to establish clear lines of responsibility and maintain public trust in AI technologies.

**Acknowledgments**. We are grateful to Killian McLoughlin and M. J. Crockett for their contributions to both the original (Earp et al., 2021) and updated (Earp et al., 2025) versions of the Relational Norms model along with some of the present authors. Many helpful comments on this article were offered by participants in the Workshop on Partiality, Relationships, and AI held at LMU Munich (24 & 25 October, 2024) organized by Ben Lange, Tom Douglas, and Sven Nyholm. Thank you also to Rodrigo Diaz for helpful comments on an earlier draft. Please note that any use of generative AI in this manuscript adheres to ethical guidelines for use and acknowledgement of generative AI in academic research (Porsdam Mann, Vazirani et al., 2024). Each author has made a substantial contribution to the work, which has been thoroughly vetted for accuracy, and assumes responsibility for the integrity of their contributions.